\documentclass{llncs}

\newif\ifdraft
\drafttrue 

\usepackage{graphicx}
\usepackage{tikz}
\usepackage{comment}
\usepackage{amsmath,amssymb} 
\usepackage{color}
\usepackage[width=122mm,left=12mm,paperwidth=146mm,height=193mm,top=12mm,paperheight=217mm]{geometry}
\usepackage{tabularx}
\usepackage{booktabs}
\usepackage{multirow}
\usepackage[normalem]{ulem}
\usepackage{caption}
\usepackage[skip=5pt]{subcaption}
\newcommand{\RR}{\mathbb{R}}
\usepackage{bm}

\newcommand{\bbold}{\bm{b}}

\newcommand{\ubold}{\bm{u}}
\newcommand{\vbold}{\bm{v}}
\newcommand{\zbold}{\bm{z}}
\newcommand{\Wbold}{\mathbf{W}}

\newcommand{\norm}[1]{\left\lVert #1 \right\rVert}
\newcommand{\T}{\mathrm{T}}

\begin{document}
\pagestyle{headings}
\mainmatter

\title{Unsupervised clustering of Roman pottery profiles from their SSAE representation
}

\ifdraft
\titlerunning{Unsuperv.\ cluster.\ of Roman pottery profiles from their SSAE represent.}
\author{Simone Parisotto\inst{1}\orcidID{0000-0003-0865-0289} \and
Alessandro Launaro\inst{2}\orcidID{0000-0002-1770-2485} \and
Ninetta Leone\inst{2}\orcidID{0000-0003-4364-5661}
\and
Carola-Bibiane Schönlieb\inst{1}\orcidID{0000-0003-0099-6306}}
\authorrunning{S. Parisotto et al.}
\institute{Centre for Mathematical Sciences, Wilberforce Road, Cambridge CB3 0WA 
\email{\{sp751,cbs31\}@cam.ac.uk}
\and
Faculty of Classics, Sidgwick Avenue, Cambridge CB3 9DA\\
\email{\{al506,nl343\}@cam.ac.uk}}
\else
\titlerunning{ECCV-20 submission ID \ECCVSubNumber} 
\authorrunning{ECCV-20 submission ID \ECCVSubNumber} 
\author{Anonymous ECCV submission}
\institute{Paper ID \ECCVSubNumber}
\fi

\maketitle

\begin{abstract}
In this paper we introduce the ROman COmmonware POTtery (ROCOPOT) database, which comprises of more than 2000 black and white imaging profiles of pottery shapes extracted from 11 Roman catalogues and related to different excavation sites. 
The partiality and the handcrafted variance of the shape fragments within this new database make their unsupervised clustering a very challenging problem: profile similarities are thus explored via the hierarchical clustering of non-linear features learned in the latent representation space of a stacked sparse autoencoder (SSAE) network, unveiling new profile matches.
Results are commented both from a mathematical and archaeological perspective so as to unlock new research directions in the respective communities.
\end{abstract}

\section{Introduction}
Our ability to \emph{interpret} an archaeological site as belonging to a specific period rests on our capacity to \emph{assign} a chronology to the material culture found in it, ranging from fixed structures to movable objects. Thanks to their relative resilience against decay combined with their specific underlying patterns of production, distribution and consumption, ceramic \emph{vessels} (pottery) – whether fragmented (i.e.\ potsherds) or intact – represent some of the most common finds recovered during archaeological fieldwork. Individual pots and potsherds are usually recorded as 2D profiles and, in consideration of their morphological features (e.g. shape of the rim or the base), are then gathered in systematic catalogues (\emph{corpora}), where patterns of similarity are used to establish relationships, in terms of function, chronology or both \cite{orton_hughes_2013}.

Within Roman archaeology specifically (albeit not exclusively), fundamental significance has been ascribed to those ceramic classes more closely associated with long-distance trade and contact, namely trade containers (\emph{amphorae}) and high-quality tableware pots (\emph{fineware}). However, these were middle-range commodities that did not reach all levels of society in the same way, and, furthermore, their supply varied enormously over time and space. Indeed, some sites might have had limited or no access to this range of objects, and such notable absence in the archaeological record might be taken to indicate their abandonment at a time when – in fact - they were in full occupation (i.e.\ reduced archaeological visibility).
In contrast, ordinary table- and kitchen-ware (\emph{commonware}) were considerably cheaper and mainly supplied within a local/regional network of distribution. As a result, they almost invariably constitute the bulk of pottery finds at almost every Roman site. Even though one would expect them to provide a most effective baseline for the dating (and more general interpretation) of Roman sites anywhere, their huge range of forms, poorly defined chronologies and scattered provenance have made their study so challenging (and so little promising) that many have favoured the analysis of the far more standardised and easily recognisable fineware and amphorae.

\begin{figure}[htb]
    \centering
    \includegraphics[width=1\textwidth,trim=5.25cm 5cm 4.75cm 6.25cm,clip=true]{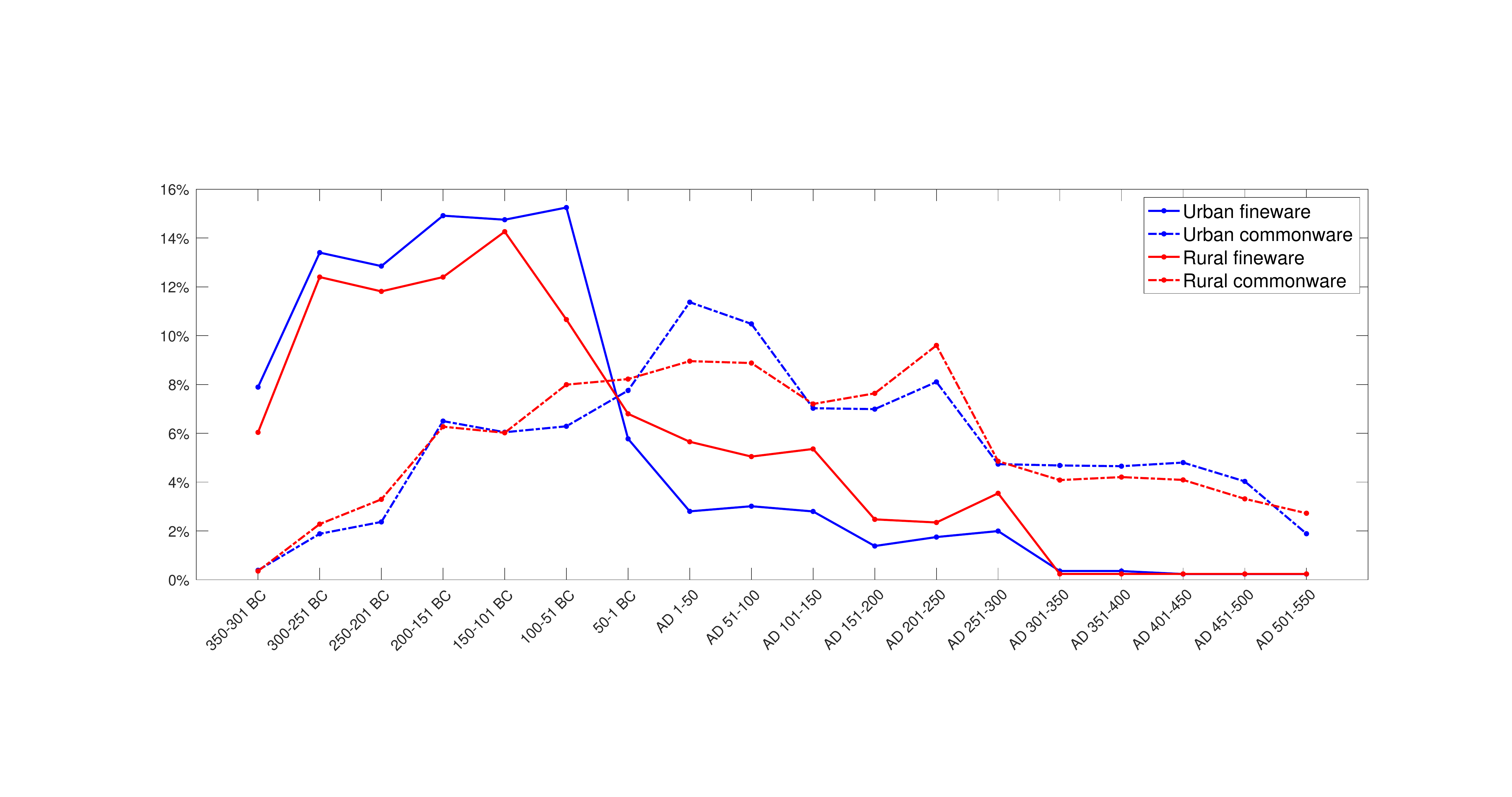}
    \caption{Percentage distribution of fineware/amphorae (continuous line) and commonware (dashed line) potsherds in urban (blue) and rural (red) environment.}
    \label{fig: fineware vs commonware}
\end{figure}

Nevertheless, when a special effort is made to include a comprehensive study of commonware sherds, resulting interpretations can dramatically change. This was neatly shown by a recent analysis of the chronological distribution of potsherds recovered from the Roman town of Interamna Lirenas and from across its surrounding countryside \cite{LauLeo2018}. Whereas traditional reliance on fineware and amphorae had outlined a trajectory of precocious demographic decline (already in progress by the late 1st century BC), the widespread presence of commonware shows a considerably more gradual process of growth, in fact peaking in the course of the 1st century AD, with little or no sign of decline until about two centuries later, see Figure \ref{fig: fineware vs commonware}. 
There can be no doubt that the study of commonware should be a priority within (Roman) material culture studies. However, this does not make the obstacles which archaeologists have to face any less real. What is indeed needed is to organise and classify this vast and varied body of evidence in a way which is considerably more effective and less time-consuming than it currently is.

\begin{figure}[htb]
    \centering
    \includegraphics[width=1\textwidth]{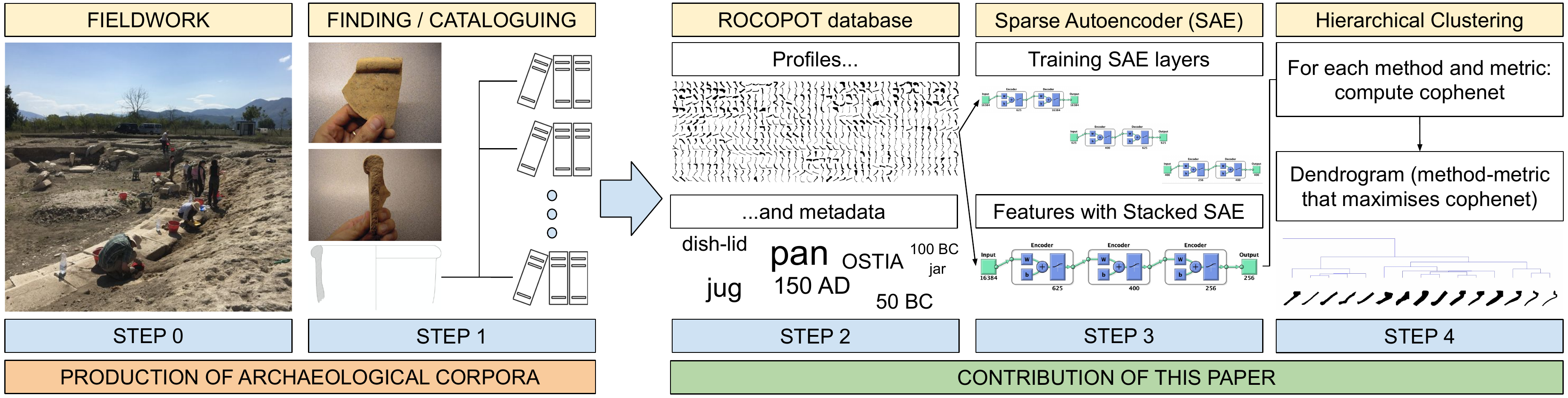}
    \caption{Summary of our contributions from the production of archaeological corpora.}
    \label{fig: workflow}
\end{figure}

\paragraph{Contributions.}
The contribution of this paper is two-fold. 
Firstly, we detail about the creation of a database of Roman commonware pottery profiles from Central Tyrrhenian Italy, called \emph{ROman COmmonware POTtery} (\texttt{ROCOPOT}) and containing 2D imaging profiles (in black and white) with metadata extrapolated from different catalogues and excavation sites. 
Secondly, we propose an unsupervised machine learning workflow for hierarchically clustering the pottery profiles by learning their features in the latent space of stacked sparse autoencoders. Our approach can thus unveil different layers of granularity and permits to discover unexpected subclasses since no comprehensive classification or ground truth is existing on these shapes. 
Our stated aim is to improve the classification performed by archaeologists on available corpora separately by effectively combining them into one all-encompassing database, with the resulting clustering patterns illuminating possible new morphological and chronological relationships and serving as an additional tool for archaeologists to improve their understanding of the chronology, function and development of both pots and sites over time. 
Whereas other similar projects have attempted to develop automated procedures for the correct identification of potsherds (i.e.\ establishing a relationship between what is newly found and the published 2D profiles), our aim is to bring an order into the database itself that facilitates the archaeological classification workflow. 
The complete pipeline, from archaeological fieldwork to our contributions, is reported in Figure \ref{fig: workflow}.
\vspace{-1em}
\vspace{-2em}
\paragraph{Related works.}
Shape recognition, classification and matching, both with complete and partial data, are well-established problems in the computer vision and mathematical communities but for settled databases of thousands of images or shapes, e.g.\ \cite{Bronstein2009,MNIST,Shilane:2004:TPS}.
In the recent years, ``Cultural Heritage Imaging Science'' became a research field of increasing popularity bringing together experts from museum institutions, history of art, physics, chemists, computer vision and mathematics departments for tackling cross-discipline challenging applications;
however, few dedicated databases emerged in the literature, mainly related to collection of paintings and associated tasks, e.g.\ for object or people detection \cite{CroZis2015,GinHasBroMal2015,Gonthier18}, style recognition \cite{BarLevWol2015} and many more, see the survey in \cite{FioKhoPonTraDBue2020}. 

Relevant works for matching 2D shapes are based on similarities of boundaries \cite{CuiFemHuWonRaz2009,SunSup2005} or skeletons \cite{ShenWangBaiWangJan2013}, homeomorphic transformations based on size functions \cite{dAmFroLan2006}, geodesic calculus for image morphing \cite{RumWir2013}, axiomatic criteria for deformable shapes \cite{Bronstein2007} or comparison of invariant image moments \cite{KimKim2000,ManCreByuYezSoa2006,Hu1962,ZunZun2014}. 
In particular, invariant shape descriptors are detailed in \cite{Cao2008} for clustering similar shapes together while still it remains open the question about the meaningfulness of the obtained clusters. 
For 3D surfaces (with triangular meshes) or volumes (with voxels), remarkable results are also obtained via scale invariant descriptors based on the eigen-decomposition of the Laplace--Beltrami operator, e.g.\ the \emph{Heat Kernel Signature} and its variants \cite{LitRodBroBro2017,RavBroBroKim2010,SunOvsGui2009}.

For the targeted application of this paper,
even fewer databases are public available \cite{CeramAlex,Tyers:1996,university_of_southampton_roman_2014}, with focus on their automatic digitisation, shape extraction and visual presentation \cite{ArchAIDE} or the clustering of a-priori selected geometric shape features into similar classes, e.g.\ by comparing curve skeletons \cite{PiccoliChiara2015Ttac}, shape boundaries \cite{Smith2012}, shape descriptors \cite{SeiWieZepPinBre2015} or employing Generalised Hough Transform distance measures (in the case of petroglyphs, similar to our images) \cite{Zhu2010} and other a-priori topological features (e.g.\ pixels height and width, area, circularity, rectangularity, diameters or steepness indexes) \cite{Christmas2018,HorrLindBrun2014} or comparison with template primitives \cite{Kampel01classificationof}. 
In \cite{BanDelEvaGatItkZal2017} fragments and nodal points are also extracted from complete 3D models in view of a \emph{supervised} deep learning approach starting from complete profiles.
In contrast, our approach for the feature extraction is totally \emph{unsupervised} and based on the latent representation space of stacked sparse autoencoders \cite{ng2011sparse}: this approach is motivated by the high availability of just fragments and the necessity to not be biased on features manually selected.

In pattern recognition, hierarchical clustering is a non-parametric yet versatile unsupervised approach for unveiling inherent structures in data or observed features, ordered in a tree called \emph{dendrogram} \cite{DudHarSto2000}. The method is based on the recursive partitioning of the data into clusters of (increasing or decreasing) cardinality, based on the minimisation of a certain cost function that promotes the separability of the data features \cite{CohKanMalMat2019}.
Despite hierarchical clustering methods are easy to implement, they are often tuned to the application at hand, with an external evaluation of the results by the experts in the applied field \cite{VekslerCourse2004}.
Specifically to Roman pottery profiles, our approach for an \emph{unsupervised hierarchical clustering} is in line with the promising works in \cite{HorrLindBrun2014,KarSmi2011,Smith2012,Zhu2010}, where, on top of that, we automate the cluster merging rule based on the best cophenetic coefficient score \cite{SokRoh1962}.
Finally, we leave the coherency check of our clusters to specialist archaeologists.

\paragraph{Organisation of the paper.} 
The paper is organised as follows:
in Section \ref{sec: database} we describe the creation of the \emph{ROman COmmonware POTtery} (\texttt{ROCOPOT}) database; 
in Section \ref{sec: problem} we detail about the tools used for clustering our shapes by means of hierarchical clustering of the learned shape features with a stacked sparse autoencoder (SSAE) network;
in Section \ref{sec: results} we present and discuss our results.

\section{The ROCOPOT Database} \label{sec: database}
The \emph{ROman COmmoware POTtery} (\texttt{ROCOPOT}) database\footnote{
The database is available at 
\ifdraft
\url{http://mach.maths.cam.ac.uk/ROCOPOT/}
\else
\{url not disclosed in the review copy\}
\fi
} described in this paper
comprises of 2475 black and white images related to shape profiles extracted from fragments of commonware Roman pottery, as catalogued in a series of \emph{corpora} \cite{DUN64,DUN65,POHL70,DYS76,CT84,CM91,ROB97,OSTIA1,OSTIA2,OSTIA3,OSTIA4}. In this work we present the version 1.0 of our database as future extensions are planned.
Since the original profiles can be composed of multiple intact parts, the specialists in our project refined the database identifying a total of 231 Bases, 278 Handles, 2103 Rims and 248 Rims with Handles. 
All of these profiles are black and white images, scanned at 300dpi from the original catalogue and saved in the lossless \texttt{.png} format with an image identification string name of the form \texttt{IDCAT-PAGNUM.FIGID.png}, where \texttt{IDCAT}, \texttt{PAGENUM} pectively, the abbreviation of the catalogue, the page number and the name of the figure where the profile appears. 
This labelling choice permits a fast inspection of the clustering results by just looking at the filenames as  similar profiles from each site tend to be grouped together within the individual catalogues. 
The details of our database are summarised in Table \ref{tab: database 1.0}, highlighting the  \texttt{IDCAT}, the bibliographic reference, the publication year, the chronology and the archaological site, as well as the number of available profiles: (O)riginals, (B)ases, (H)andles, (R)ims and Rims with Handles (RW).

\begin{figure}[htb]
\centering
\begin{subfigure}[t]{0.32\textwidth}\centering
\includegraphics[height=0.78cm]{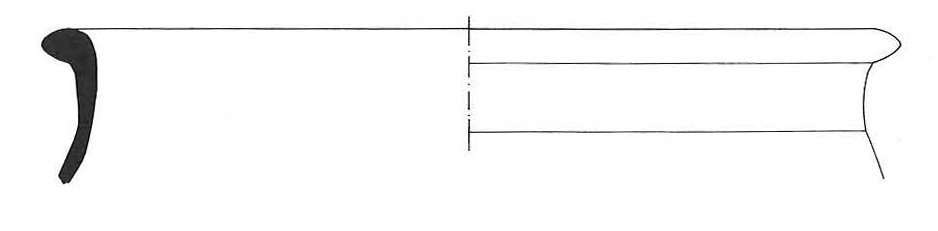}
\includegraphics[height=0.78cm]{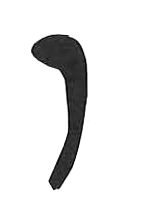}
\captionsetup{justification=centering}
\caption{DYS76-3.CF39}
\label{fig: rim separation}
\end{subfigure}
\begin{subfigure}[t]{0.32\textwidth}\centering
\includegraphics[height=0.78cm]{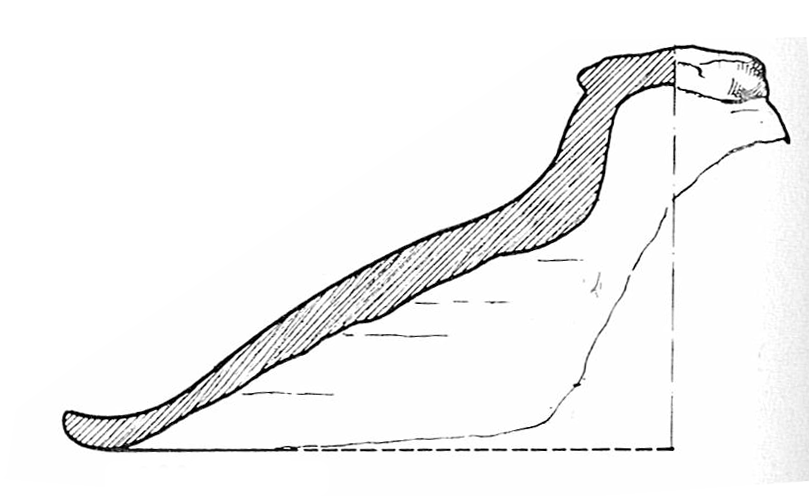}
\includegraphics[height=0.78cm]{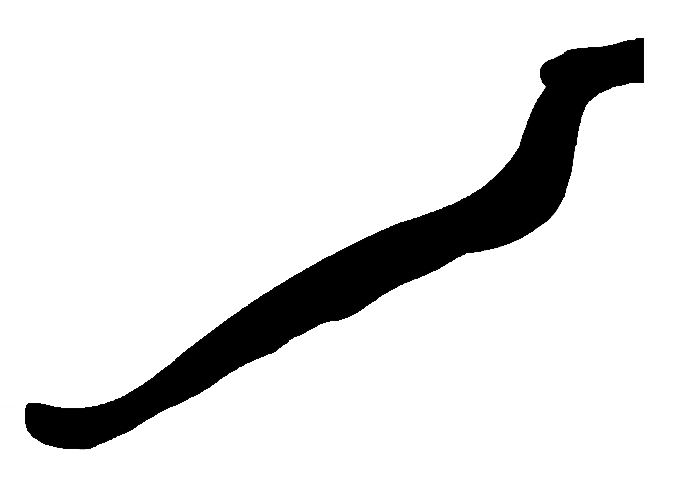}
\captionsetup{justification=centering}
\caption{POHL70-55.38}
\label{fig: rim filling 1}
\end{subfigure}
\begin{subfigure}[t]{0.32\textwidth}\centering
\includegraphics[height=0.78cm]{\detokenize{./images/preprocessing/CT84_original}.png}
\includegraphics[height=0.78cm]{\detokenize{./images/preprocessing/CT84}.png}
\captionsetup{justification=centering}
\caption{CT84-99.1 CE 2083 3a}
\label{fig: rim filling 2}
\end{subfigure}
\\
\begin{subfigure}[t]{0.32\textwidth}\centering
\includegraphics[height=0.78cm,trim=1em 3.1em 1em 3.5em,clip=true]{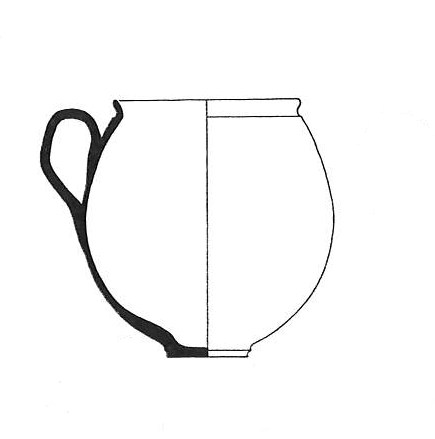}
\includegraphics[height=0.78cm]{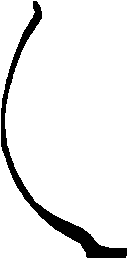}
\captionsetup{justification=centering}
\caption{DUN64-7.3}
\label{fig: rim polishing 1}
\end{subfigure}
\begin{subfigure}[t]{0.32\textwidth}\centering
\includegraphics[height=0.78cm]{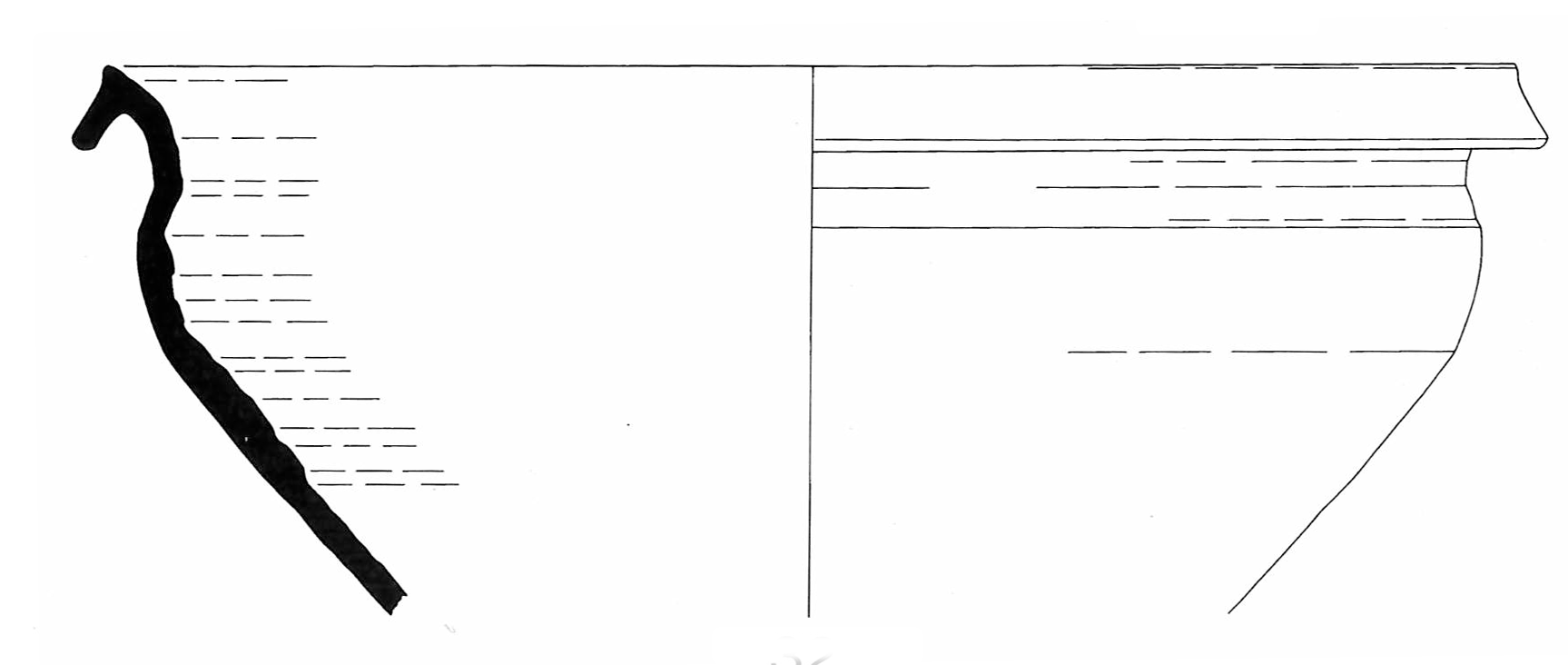}
\includegraphics[height=0.78cm]{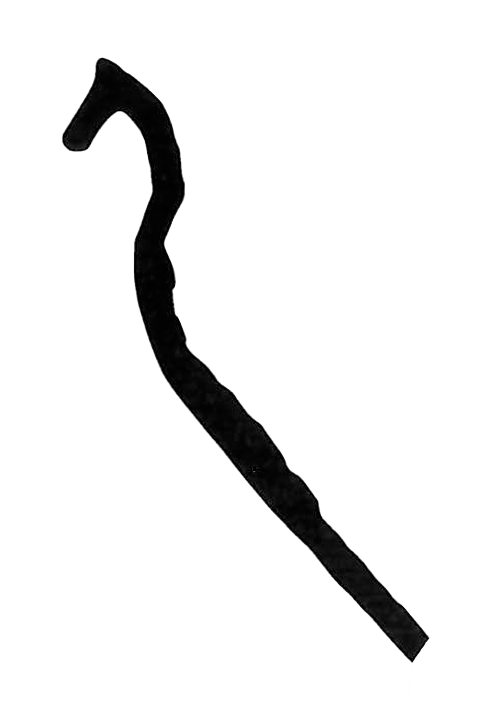}
\captionsetup{justification=centering}
\caption{ROB97-220.32}
\label{fig: rim polishing 2}
\end{subfigure}
\begin{subfigure}[t]{0.32\textwidth}\centering
\includegraphics[height=1.3cm,trim=0em 1cm 0em 0em,clip=true]{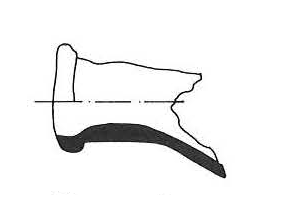}
\includegraphics[height=0.7cm]{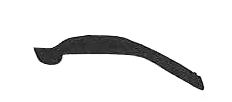}
\captionsetup{justification=centering}
\caption{DYS76-17.16IV69}
\label{fig: DYS76-17.16IV69}
\end{subfigure}
\caption{Shape preprocessing. Subfigures: original (left) and extracted (right) shape.}
\label{fig: preprocessing}
    \begin{subfigure}[t]{0.16\textwidth}\centering
    \includegraphics[height=2cm]{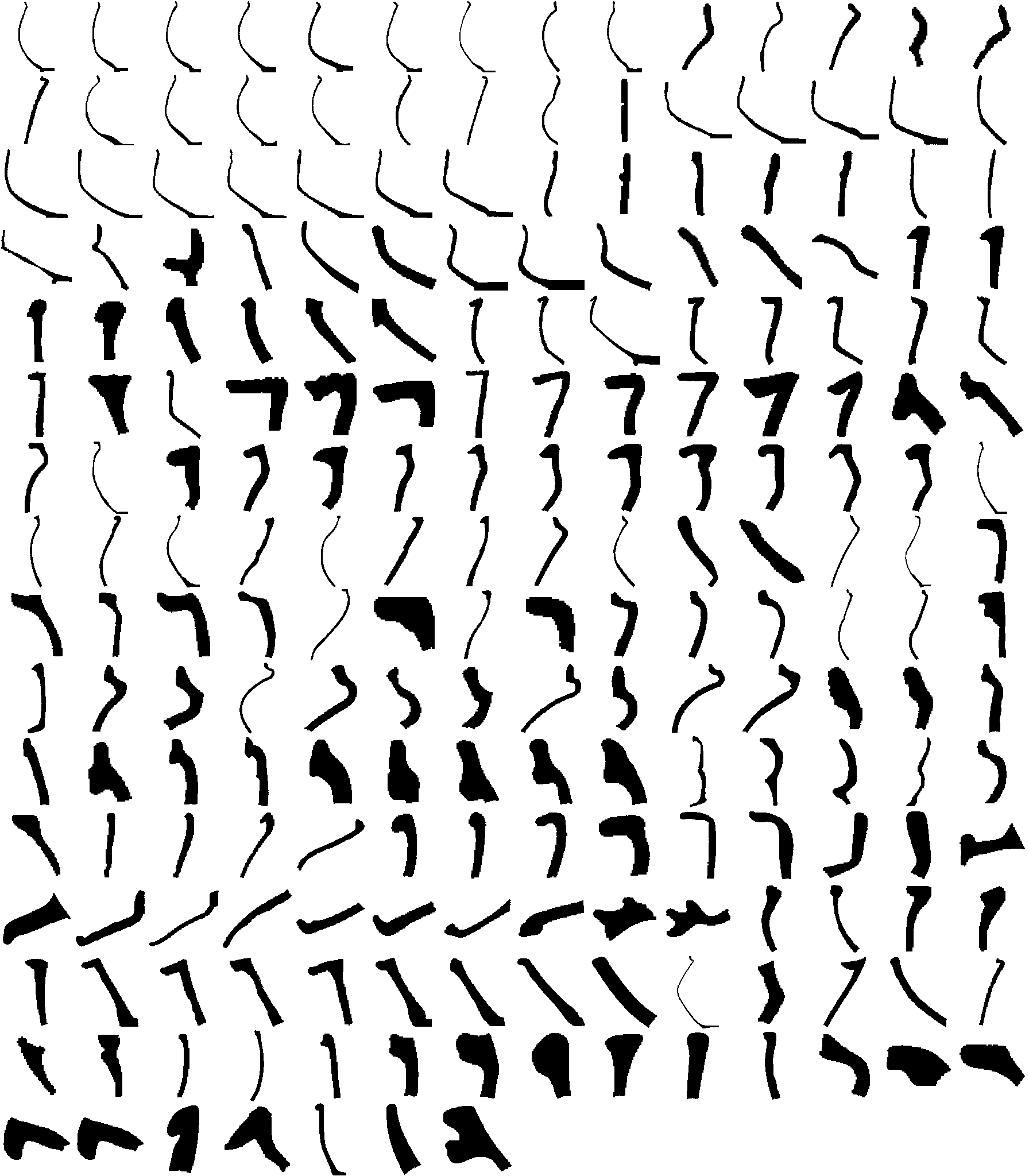}
    \caption{DUN64}
    \end{subfigure}
    \hfill
    \begin{subfigure}[t]{0.16\textwidth}\centering
    \includegraphics[height=2cm]{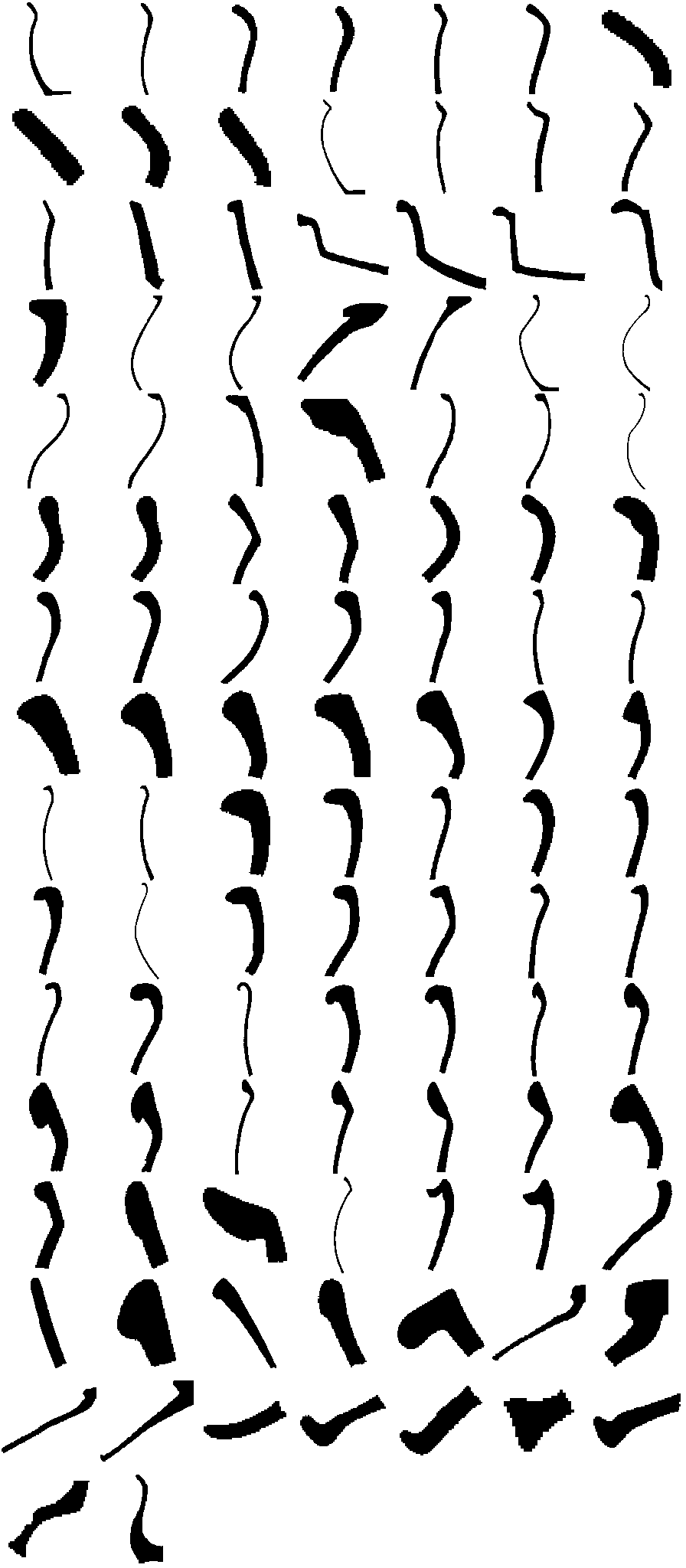}
    \caption{DUN65}
    \end{subfigure}
    \hfill
    \begin{subfigure}[t]{0.16\textwidth}\centering
    \includegraphics[height=2cm]{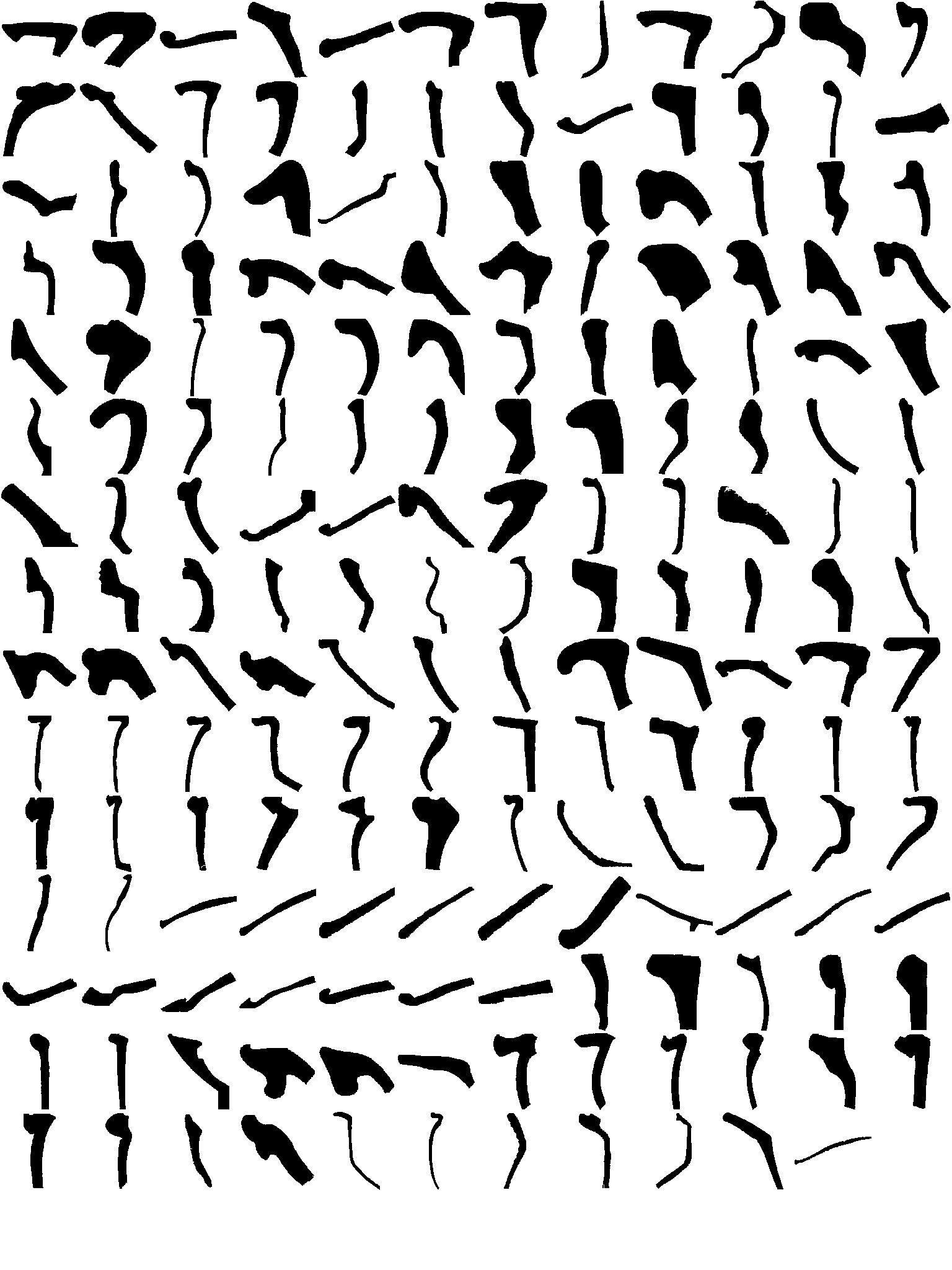}
    \caption{POHL70}
    \end{subfigure}
    \hfill
    \begin{subfigure}[t]{0.48\textwidth}\centering
    \includegraphics[height=2cm]{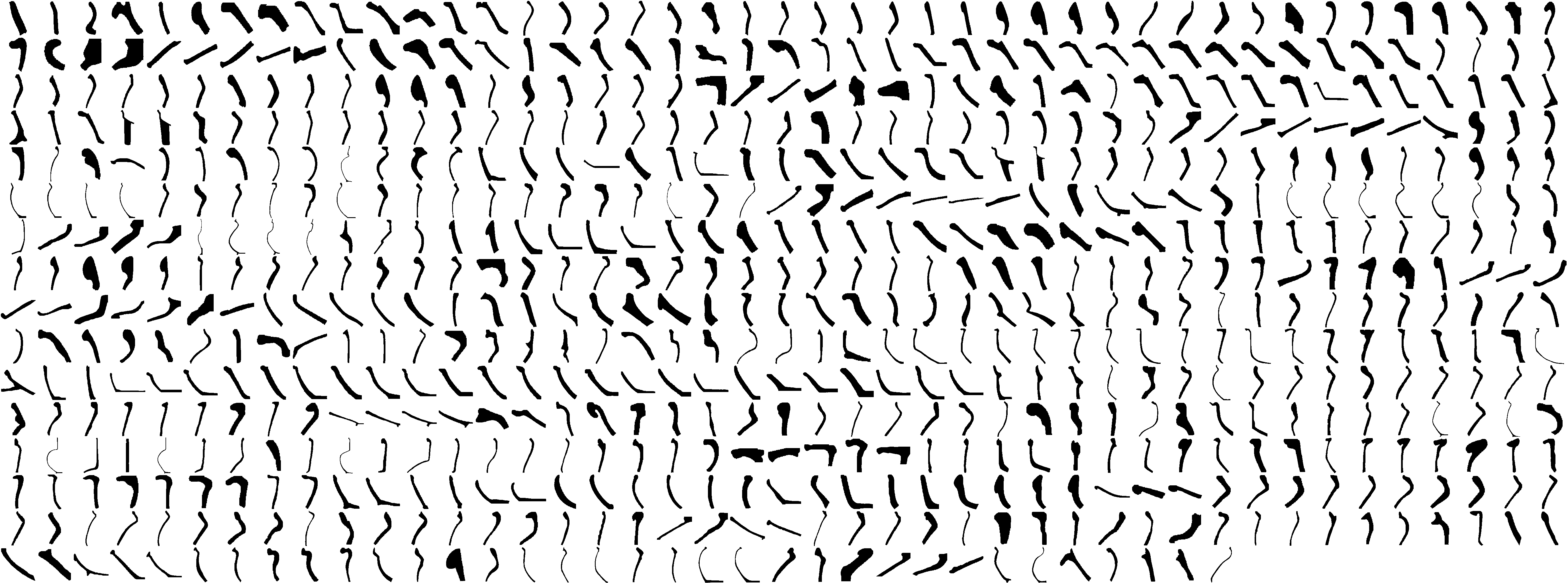}
    \caption{DYS76}
    \end{subfigure}
    \\
    \begin{subfigure}[t]{0.135\textwidth}\centering
    \includegraphics[height=2cm]{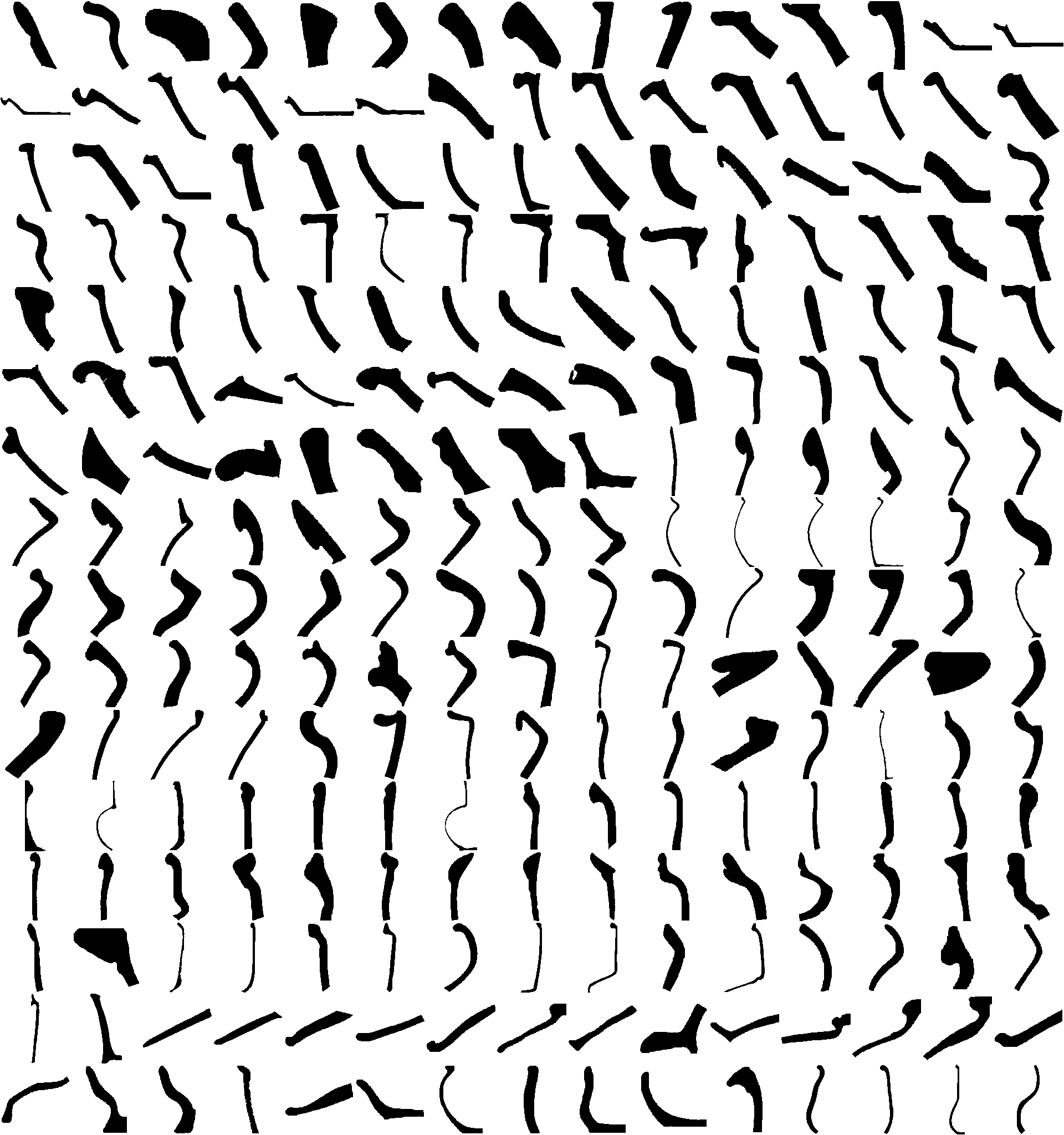}
    \caption{CT84}
    \end{subfigure}
    \begin{subfigure}[t]{0.135\textwidth}\centering
    \includegraphics[height=2cm]{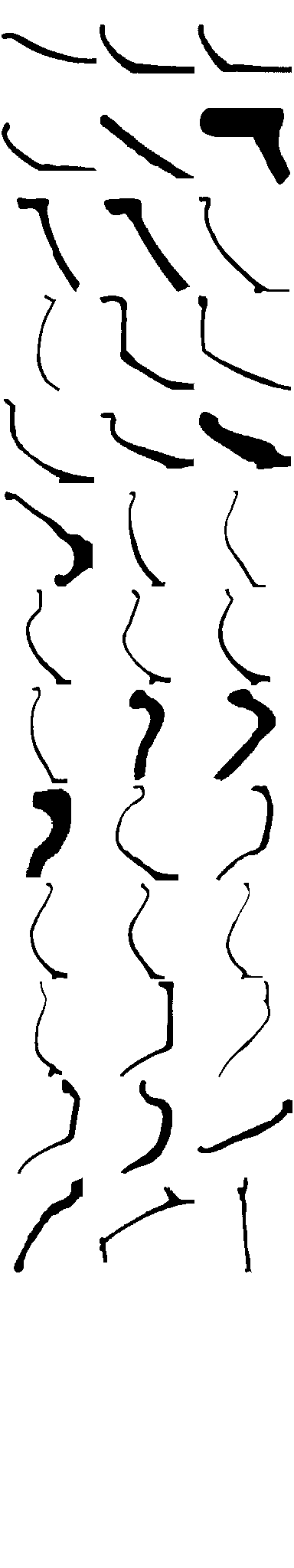}
    \caption{CM91}
    \end{subfigure}
    \begin{subfigure}[t]{0.135\textwidth}\centering
    \includegraphics[height=2cm]{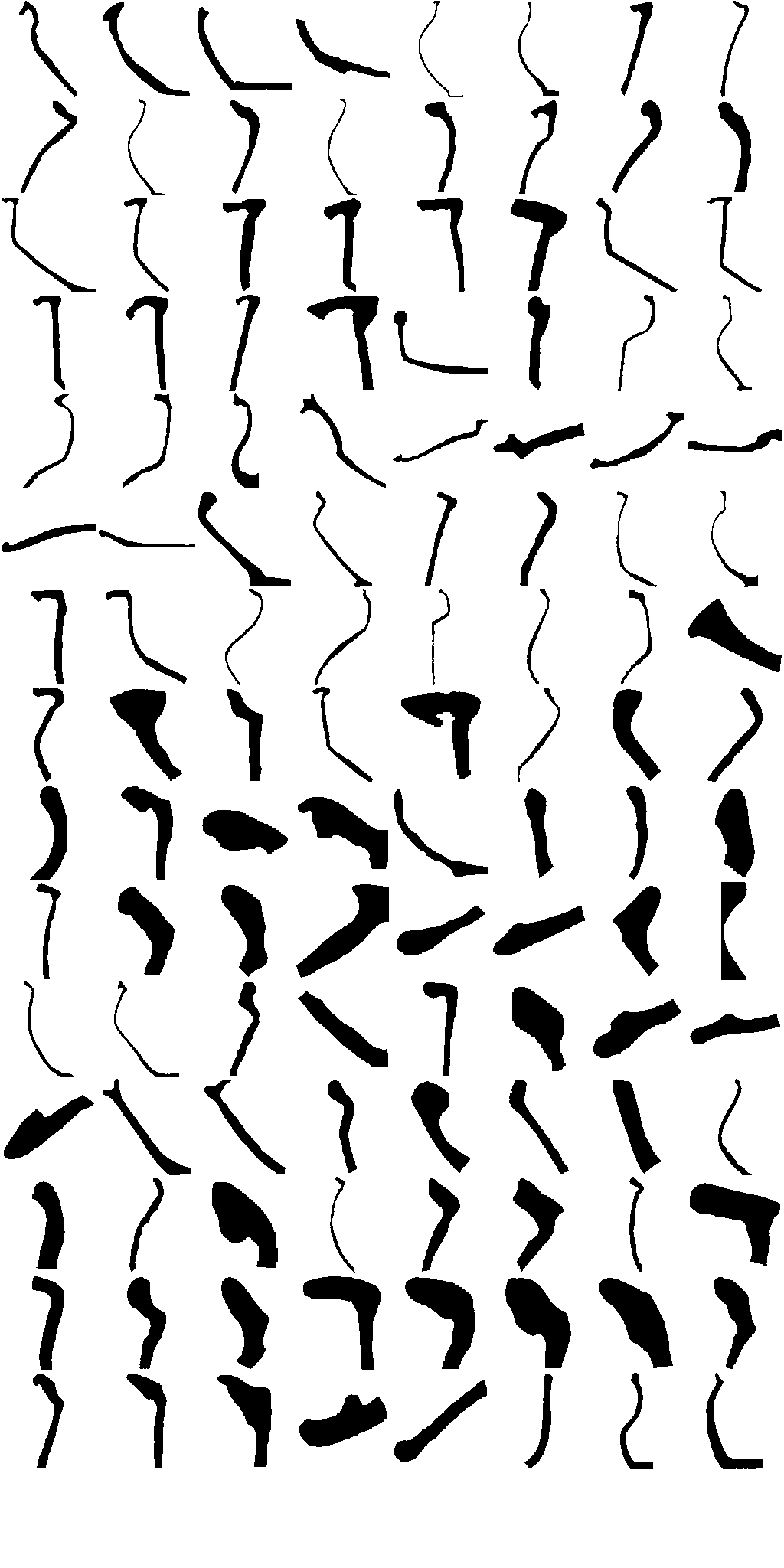}
    \caption{ROB97}
    \end{subfigure}
    \begin{subfigure}[t]{0.135\textwidth}\centering
    \includegraphics[height=2cm]{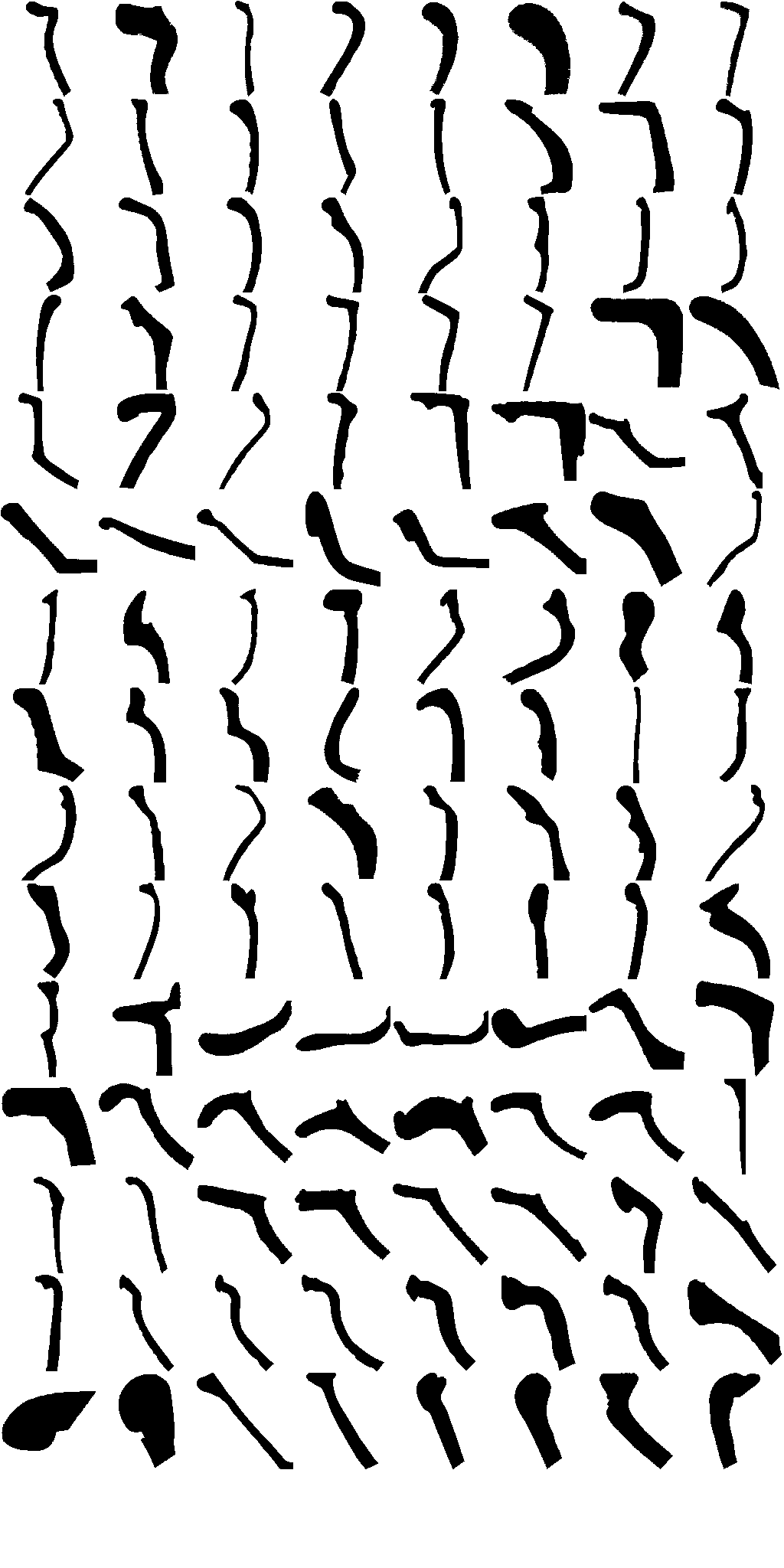}
    \caption{OSTIA1}
    \end{subfigure}
    \begin{subfigure}[t]{0.135\textwidth}\centering
    \includegraphics[height=2cm]{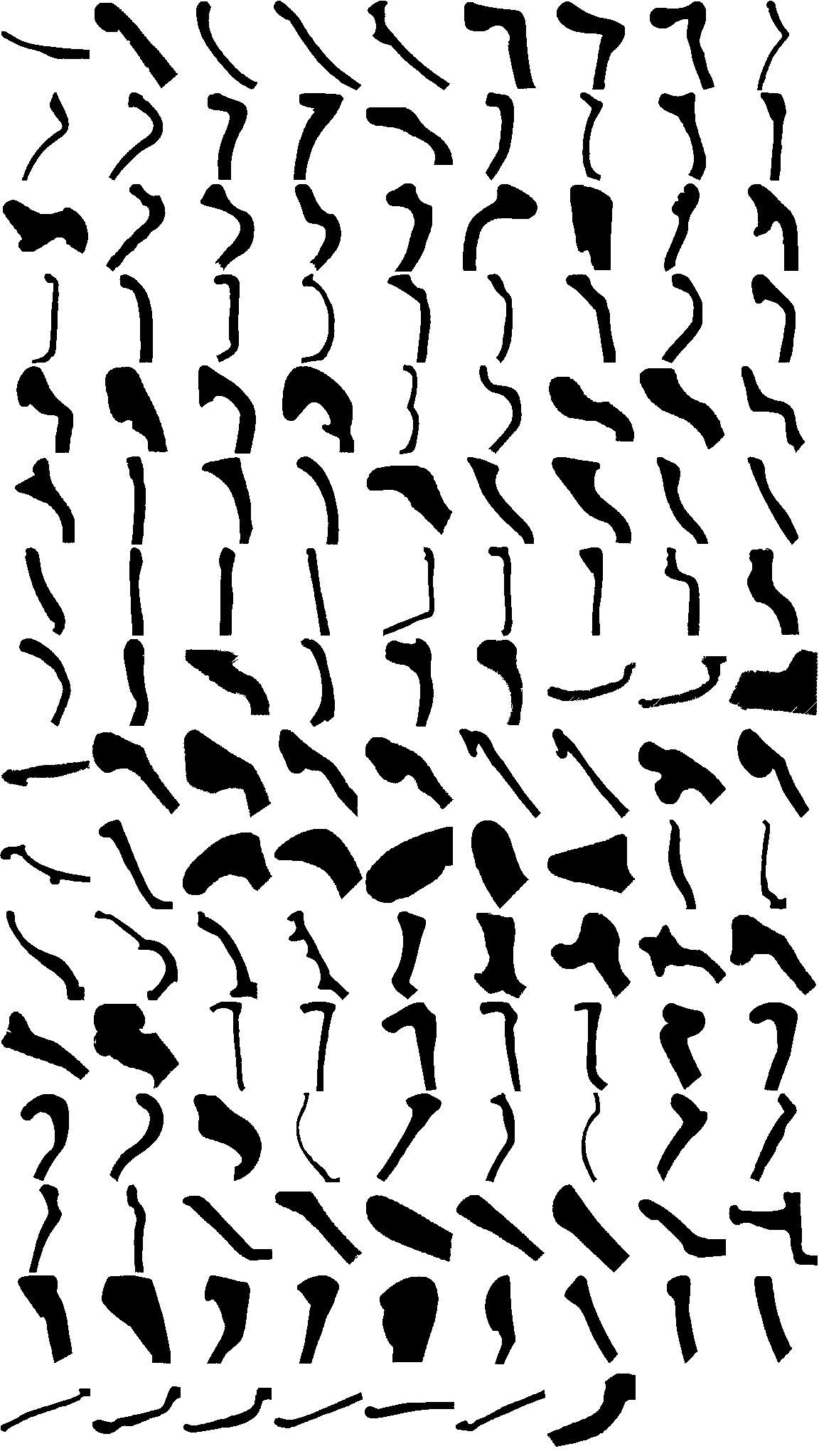}
    \caption{OSTIA2}
    \end{subfigure}
    \begin{subfigure}[t]{0.135\textwidth}\centering
    \includegraphics[height=2cm]{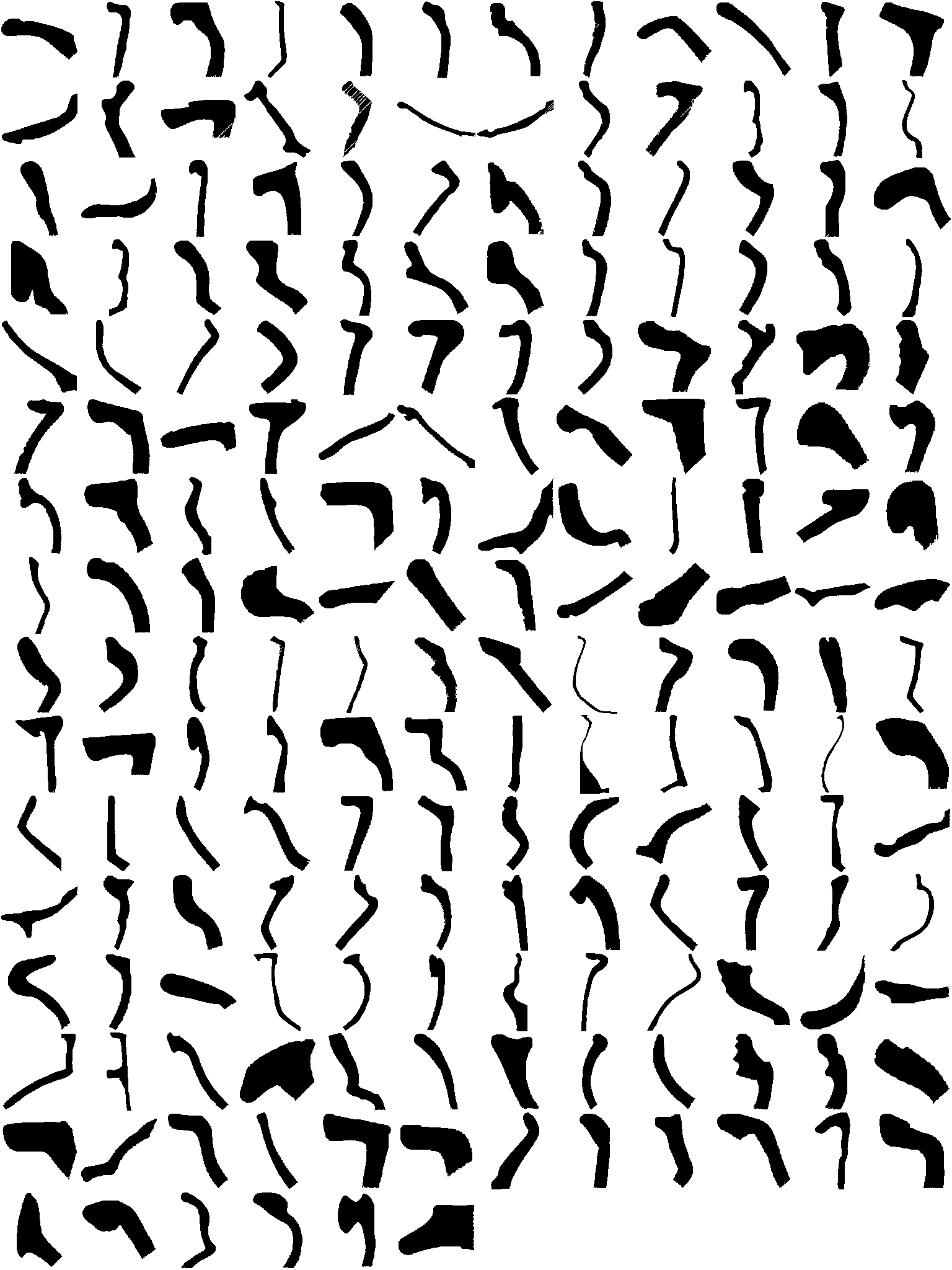}
    \caption{OSTIA3}
    \end{subfigure}
    \begin{subfigure}[t]{0.135\textwidth}\centering
    \includegraphics[height=2cm]{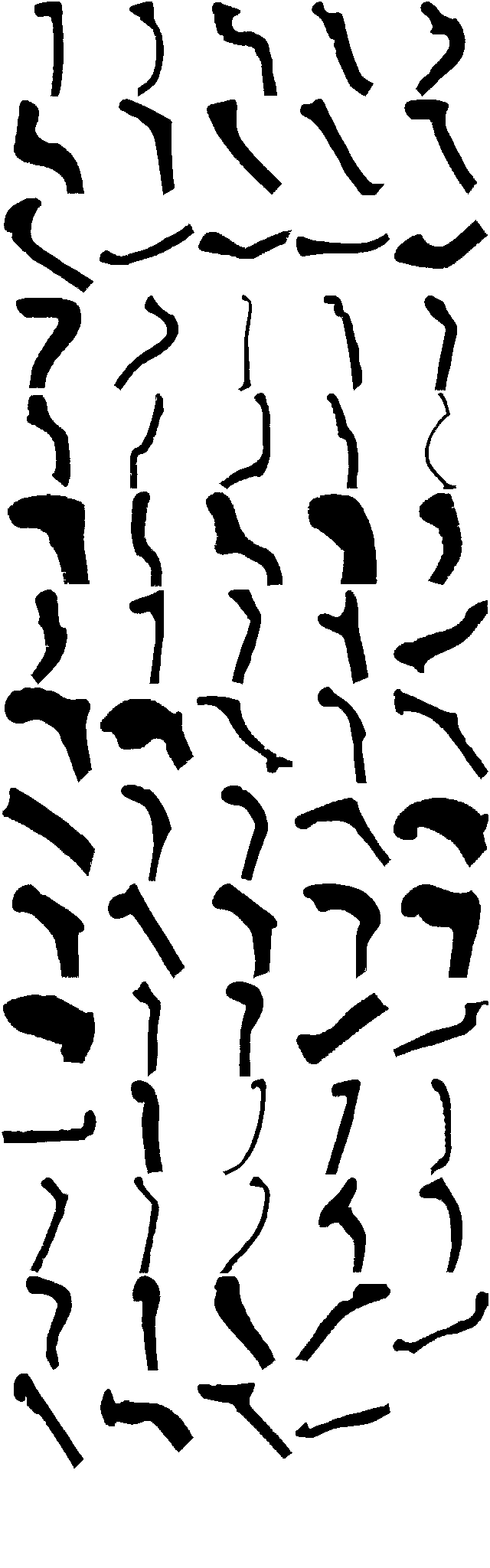}
    \caption{OSTIA4}
    \end{subfigure}
    \caption{Roman Pottery database v1.0: catalogues with (R)ims shape profiles.}
    \label{fig: database 1.0}
\end{figure}

However, all the extracted profiles require a further polishing step so as to make them uniformly represented in the image space before the feature extraction. As a matter of example, in Figure \ref{fig: preprocessing} we report common situations when extracting the ``Rims'', e.g.\ the separation of the rim from the template (Figure \ref{fig: rim separation}), the fill in of relevant portions with black colour intensity (Figure \ref{fig: rim filling 1} and \ref{fig: rim filling 2}), the cleaning of scanned contours from ageing phenomena and the removal of undesired handles (Figure \ref{fig: rim polishing 1} and \ref{fig: rim polishing 2}).
The expertise of specialist archaeologists is of vital importance at this stage so as to prevent incorrect digitisation of profiles. As an example, the case in Figure \ref{fig: DYS76-17.16IV69} provides a clear situation where the horizontal features plays a role: if such shape had been rotated clockwise by $90^\circ$ degrees, than it would have matched with a different cluster.

Finally, we make the different profiles comparable by resizing them to a size of 128$\times$128 pixels, without distorting the aspect ratio, see Figure \ref{fig: database 1.0}: this normalisation is essential in view of the feature extraction with sparse autoencoders.

\begin{table}[htb]\scriptsize
\caption{Details of the Roman Pottery database v1.0, with number of (O)riginal shapes and their (B)ases, (H)andles, (R)ims and Rims with Handles (RH).}
\label{tab: database 1.0}
\begin{tabularx}{1\textwidth}{Xccc|c|c|c|ccc}
\toprule
\texttt{IDCAT} & \multirow{2}{*}{Ref.} & \multirow{2}{*}{Year} & \multicolumn{5}{c}{Shapes} & \multirow{2}{*}{Chronological range} & \multirow{2}{*}{Site} \\
\cmidrule(r){4-8}
(catalogue abbreviation) &&& O & B & H & R & RH &&\\
\midrule
DUN64  & \cite{DUN64}  & 1964 & 245 & 13 &  35 &  217 &  37 & 50 BC - 26 BC & Sutri\\
DUN65  & \cite{DUN65}  & 1965 & 127 &  13 &  11 &  107 &   9 & 150 BC - 1 BC & Sutri\\
POHL70 & \cite{POHL70} & 1970 & 182 &   3 &   9 &  179 &  22 & 200 BC - AD 140 & Ostia\\
DYS76  & \cite{DYS76}  & 1976 & 814 & 109 &  62 &  679 &  46 & 350 BC - AD 335 & Cosa\\
CT84   & \cite{CT84}   & 1984 & 269 &  28 &  26 &  240 &  14 & 700 BC - AD 100 & Pompei\\
CM91   & \cite{CM91}   & 1991 & 43 &  4 &  10 &   39 &   4 & 100 BC - AD 200 & La Celsa\\
ROB97  & \cite{ROB97}  & 1997 & 132 & 10 &  21 &  120 &  13 & AD 101 - AD 635 & Monte Gelato\\
OSTIA1 & \cite{OSTIA1} & 1968    & 168 &  26 &  45 &  120 &  42 & AD 101 - AD 500 & Ostia\\
OSTIA2 & \cite{OSTIA2} & 1970    & 175 &  20 &  19 &  142 &  24 & AD 51 - AD 150 & Ostia\\
OSTIA3 & \cite{OSTIA3} & 1973    & 230 &  32 &  31 &  186 &  23 & AD 51 - AD 500 & Ostia\\
OSTIA4 & \cite{OSTIA4} & 1977 & 90 &  13 &   9 &   74 &  14 & AD 251 - AD 425 & Ostia\\
\midrule
ALL &                & &   2475  & 231 & 278 & 2103 & 248 & &\\ 
\bottomrule
\end{tabularx}
\end{table}

\section{Proposed approach}\label{sec: problem}
After the preprocessing step is carried out, we proceed with the main idea to learn the low-dimensional representation of the shape profiles in our database with sparse autoencoders (SAE) in order to hierarchically cluster the images based on a reduced number of features. 
Note that the combination of these two approaches makes our workflow totally unsupervised, a key assumption for questioning the established classification in archaeological corpora and unveiling new clustering patterns.
For the sake of completeness, we are going to detail in the next subsections the main concepts of these two tools.

\subsubsection*{Sparse Autoencoders.}
The first step is to produce a low dimensional set of features that is able to represent the shapes in the database at their best. 
Instead of fixing a-priori the quality of the shape features (e.g.\ the diameter, the ballness, the elongation, the area, the perimeter and many more), we decided to learn the shape representation in a reduced dimensional space of fixed dimension. In this sense, sparse autoencoders are powerful tools as they can be interpreted as a non-linear version of the principle component analysis.

In machine learning, a \emph{sparse autoencoder} (SAE) is an artificial neural network trained for replicating its input $\ubold\in\RR^d$ at its output $\widehat{\ubold}\in\RR^d$, with $d>0$, by means of back-propagating the reconstructed result so as to learn the input representation onto a smaller dimensional space of \emph{neurons} via the optimisation of an error measure function called \emph{loss function}. Here, the \emph{sparsity} assumption forces the neurons to specialise on few data. In other words, an autoencoder is designed to learn the identity function of the input, see \cite{ng2011sparse}. 
An autoencoder is composed by two main functions, called \emph{encoder} and \emph{decoder}. The encoder is dedicated to the dimensional reduction of the input to the latent subspace, from which the decoder reconstructs the output. 
More advanced \emph{stacked sparse autoencoders} (SSAE) \cite{VicLarLajBenMan2010} are neural networks composed of multiple SAE, where the output of the hidden layer of one autoencoder is inputted to the next autoencoder: this strategy is promising when learning high-level features \cite{Xu2016}.  In what follows, we detail the main concepts behind SSAE.

Let $\ell=0,\dots,L$ be a variable indexing the layer level up to level $L>0$ and let $(k_\ell)_{\ell=0}^L$ be a monotonic decreasing numeric sequence, with $k_0=d$ and $k_\ell$ indicating the number of neurons that can be activated at any fixed layer $\ell>0$.
For $L=1$, the \emph{single-layer encoder} network is composed by an encoder map $f^{(1)}:\RR^{k_0}\to\RR^{k_1}$ defined as $\vbold^{(1)} = f^{(1)}(\Wbold^{(1)} \ubold + \bbold_\text{e}^{(1)})$ with weights $\Wbold^{(1)}\in\RR^{k_1\times k_0}$ and a bias vector $\bbold_\text{e}^{(1)}\in\RR^{k_1}$, 
and a decoder map $g^{(1)}:\RR^{k_1}\to\RR^{k_0}$ defined as $\widehat{\ubold} = g^{(1)}(\Wbold^{(1),\T} \vbold^{(1)} + \bbold_\text{d}^{(1)})$ with weights $\Wbold^{(1),\T}\in\RR^{k_0\times k_1}$, $(\,\cdot\,)^\T$ indicates the transpose operation and bias vector $\bbold_\text{d}^{(1)}\in \RR^{k_0}$. 
For $L>1$, the {multi-layer autoencoder} network is based on the encoder-decoder strategy iterated for $L$ layers: for example, in a two layers strategy the output $\vbold^{(1)}$ of $f^{(1)}$ is concatenated with the encoder $f^{(2)}:\RR^{k_1}\to\RR^{k_2}$ via $\vbold^{(2)}=f^{(2)}(\Wbold^{(2)}\vbold^{(1)} + \bbold_\text{e}^{(2)})$, with weights $\Wbold^{(2)}\in\RR^{k_2\times k_1}$ and bias vector $\bbold_\text{e}^{(2)}\in\RR^{k_2}$, and a decoder map $g^{(2)}:\RR^{k_2}\to\RR^{k_1}$ via $\widehat{\vbold}^{(1)}=g^{(2)}(\Wbold^{(2),\T}\vbold^{(2)}+\bbold_\text{d}^{(2)})$, with weights $\Wbold^{(2),\T}\in\RR^{k_2\times k_1}$ and bias vector $\bbold_\text{d}^{(2)}\in\RR^{k_1}$.
For a generic vector $\zbold\in\RR^{k_\ell}$, popular choices for the encoder $f^{(\ell)}$ and decoder $g^{(\ell)}$ maps are any between the following transfer functions $h:\RR^{k_\ell}\to\RR^{k_\ell}$: the \texttt{logsig} $h(\zbold) = \left(1+\exp(-\zbold)\right)$; the \texttt{satlin} function defined by cases as $h(\zbold) = 0$ if $\zbold\leq 0$, $h(\zbold) = 1$ if $\zbold\geq 1$ and $h(\zbold)=\zbold$ otherwise; the \texttt{purelin} function $h(\zbold) = \zbold$.

When the number of neurons increase, then it is desiderable to keep their activation as small as possible, boosting their specialisation. In order to achieve this purpose, a key concept is the \emph{average output activation measure} $\widehat{\rho}_i$ associated to the neuron $i$. Indeed, suppose that we have $N$ training samples $(\ubold_n)_{n=1}^N$ in the database, with each $\ubold_n\in\RR^d$. For the generic transfer map $h$ (and for simplicity in the single-layer case $L=1$), then such measure is defined as:
\begin{equation}
\widehat{\rho}_i 
= 
\frac{1}{N} 
\sum_{n=1}^{N} 
h
\left( 
\prod_{j=1}^{k_0}
\left(
\Wbold^{(1)}_{i,j} (\ubold_n)_j
\right)
+ (\bbold_\text{e}^{(1)})_i
\right).
\label{eq: average output activation measure}
\end{equation}
Thus, a neuron is considered to be activated if the value of \eqref{eq: average output activation measure} is high; conversely, the neuron results sensitive to a small number of training examples and encouraged to learn a \emph{sparse} representation of the observed samples.
The sparsity goal is achievable with a \emph{sparse regulariser}, that is the Kullback-Leibler (KL) divergence and measuring the difference between distributions. In particular, we aim to find an average activation measure $\rho$ whose distribution is close to $\widehat{\rho}_i$
\begin{equation}
\mathcal{L}_\text{sparsity}
=
\sum_{i=1}^{k_0} \mathrm{KL}
\left(
\rho\parallel \widehat{\rho}_i
\right)
=
\sum_{i=1}^{k_0} 
\rho \log\left(\frac{\rho}{\widehat{\rho}_i}\right)
+
(1-\rho)\log\left(\frac{1-\rho}{1-\widehat{\rho}_i}\right).
\label{eq: loss sparsity}
\end{equation}
However, the (KL)-divergence is not sufficient by itself to achieve the goal as an increasing number of neurons would obtain a similar results \cite{OLSHAUSEN19973311}. Therefore, it is necessary to control the magnitude of the weights and the overfitting by considering  a further $\mathrm{L}_2$-regularisation term on the weights $\Wbold^{(\ell)}$:
\begin{equation}
\mathcal{L}_\text{weights} = \norm{\Wbold^{(\ell)}}_2^2
=
\sum_{\ell=1}^L 
\sum_{i=1}^{k_{\ell}}
\sum_{j=1}^{k_{\ell-1}}
\Wbold^{(\ell)}_{ij}.
\label{eq: loss weights}
\end{equation}

Finally, a mean squared error function between the input and its reconstruction is also minimised for each example in the database
\begin{equation}
\mathcal{L}_\text{mse}
=
\sum_{n=1}^N \norm{\ubold_n - \widehat{\ubold}_n}_2^2
=
\sum_{n=1}^N \sum_{j=1}^{k_0}\left((\ubold_n)_j - (\widehat{\ubold}_n)_j\right)^2.
\label{eq: loss mse}
\end{equation}

For training each layer in the stacked autoencoder, the following global loss function is a modified mean squared error function that considers all the three terms in  \eqref{eq: loss sparsity}, \eqref{eq: loss weights} and \eqref{eq: loss mse}:
\begin{equation}
\mathcal{L}_\text{total} 
= 
\beta
\mathcal{L}_\text{sparsity}
+
\lambda
\mathcal{L}_\text{weights}
+
\mathcal{L}_\text{mse}.
\label{eq: cost function}
\end{equation}

Even if sparse autoencoders are observed to reconstruct blurry data, they still offer a favourable tool for our application as we can take this weakness into an advantage to overcome the pixelisation of our black and white imaging data, which are digitised from catalogues of poor printing quality or in bad conservation state.
Moreover, despite the unsupervised approach may result into a weak performance for the general labelling problem (since we do not back-propagate any label as in a semi-supervised learning approach), it still fits our purposes to postpone any feedback to the coherency of the results and possibly unveiling new clustering patterns in catalogues.

Details about the particular network chosen for our applications are given in Section \ref{sec: results}. For the next subsection it is sufficient to focus on the latent subspace of the deepest layer of features in the stacked autoencoder network, i.e.\ $\vbold^{(L)}\in\RR^{k_L}$. 

\subsubsection*{Hierarchical clustering.} 
Once the $\{k_1,\dots,k_L\}$ neurons for encoding-decoding all the shape profiles are trained, we are now in the position to obtain for each profile a final vector of $k_L$ features, representing its specific \emph{signature} in the latent representation space of SSAE.
These features can be used for clustering the shape profiles and unveiling their similarities.
Since popular clustering methods, like \texttt{k-means}, have the bottleneck to require as input the number of expected clusters, which is unknown for our database, we opted for an \emph{agglomerative hierarchical clustering} approach, which creates a tree of clusters by starting with a class for each shape, subsequently merged together in a bottom-up hierarchical strategy according to a selected \texttt{method} for computing the distance between clusters and a selected \texttt{metric} between the observed features. 
The unsupervised selection of the best pair of \texttt{method} and \texttt{metric}, within the range of possibilities described in Table \ref{tab: metric method}, is performed by computing for each possible pair the \texttt{cophenet} correletion coefficient, which measures how faithfully the selected tree represents the dissimilarities among observations. 
In view of this task, we need the \emph{cophenetic} similarity measure for any pair of observations, from which the \emph{cophenet} is computed \cite{SokRoh1962}.
Thus, the pair with the highest \texttt{cophenet} value corresponds to the selected \texttt{method} and \texttt{metric}.
Finally, a \emph{dendrogram} tree linking all the clusters is produced, with the top-level class (the \emph{root}) containing all the shapes while the bottom-level classes (the \emph{leaves}) containing the most similar shapes. Here, we call \emph{seeds} all the leaves containing a pair of shapes, whose matching will be evaluated by specialist archaeologists in Section \ref{sec: results}.
 
\vspace{-2em}
\begin{table}[!htb]
\caption{Range of \texttt{method} and \texttt{metric} for our hierarchical clustering approach.}\scriptsize
\label{tab: metric method}
\begin{tabularx}{1\textwidth}{l|X}
\toprule
\multicolumn{2}{c}{\texttt{Method} (distance between clusters)}\\
\midrule
 \texttt{single} & Shortest distance \cite{Florek1951}\\
 \texttt{complete} & Farthest distance \cite{sorensen1948method} \\ 
 \texttt{average} & Unweighted pair group method using arithmetic averages (UPGMA) \cite{sokal58}\\ 
 \texttt{weighted} & Weighted pair group method using arithmetic averages (WPGMA) \cite{sokal58}\\ 
 \texttt{centroid} & Unweighted pair group method using centroids (UPGMC) \cite{Milligan1980}\\ 
 \texttt{median} & Weighted pair group method using centroids (WPGMC) \cite{JainDubes1988}\\ 
 \texttt{ward} & Minimum variance \cite{Ward1963}\\
\toprule
\multicolumn{2}{c}{\texttt{Metric} (distance between observed features)}\\
\midrule
\texttt{euclidean} & Euclidean distance\\
\texttt{squaredeuclidean} & Squared Euclidean distance \\
\texttt{seuclidean} & Standardized Euclidean distance (scaled based on the standard deviation)\\
\texttt{cityblock} & City-block distance\\
\texttt{minkowski} & Minkowski distance (default exponent is 2)\\
\texttt{chebychev} & Maximum coordinate difference\\
\texttt{cosine} & One minus the cosine of the included angle between points (treated as vectors)\\
\bottomrule
\end{tabularx}
\end{table}
\vspace{-2em}

\section{Results}\label{sec: results}
In this section we detail about the parameters of the stacked sparse autoencoders network and the workflow of our hierarchical clustering approach applied to the rims profiles in our database. Then we discuss our results from a mathematical and archaeological point of view. 
In what follows, we consider each row of Table \ref{tab: database 1.0} as a separate experiment with the associate number of profiles: each test is called with its \texttt{IDCAT} label (note that \texttt{IDCAT}=ALL means that all the profiles available are considered together).
All the tests are performed on a MacBook Pro 13", 2.4 GHz Intel Core i5 quad-core with 16 GB of RAM 2133 MHz LPDDR3. The software is implemented in MATLAB 2019b.

\subsubsection*{Parameters and analysis of the SSAE network.}
\begin{figure}[!htb]
    \centering
    \begin{subfigure}[t]{0.32\textwidth}\centering
    \includegraphics[width=1\textwidth,trim=0.2em 0.2em 0.2em 0.2em,clip=true]{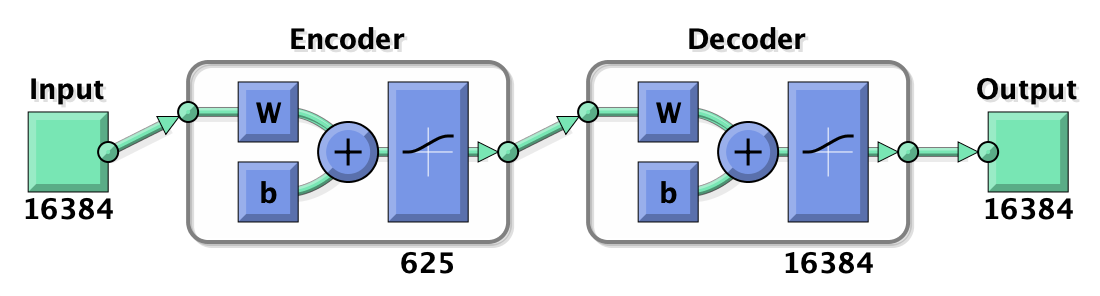}
    \caption{SAE (Layer 1)}
    \end{subfigure}
    \hfill
    \begin{subfigure}[t]{0.32\textwidth}\centering
    \includegraphics[width=1\textwidth,trim=0.2em 0.2em 0.2em 0.2em,clip=true]{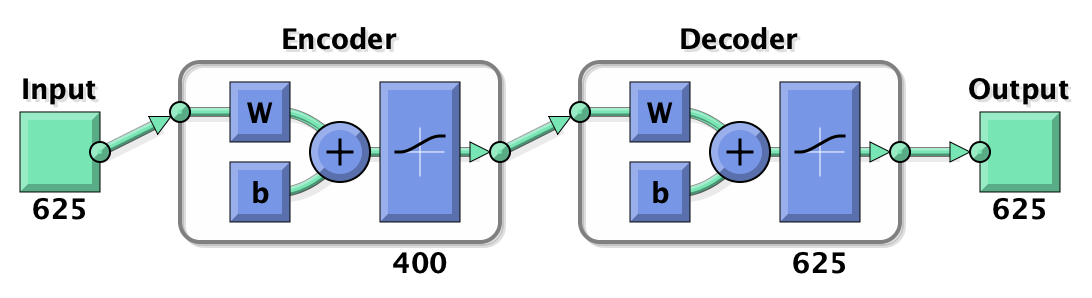}
    \caption{SAE (Layer 2)}
    \end{subfigure}
    \hfill
    \begin{subfigure}[t]{0.32\textwidth}\centering
    \includegraphics[width=1\textwidth,trim=0.2em 0.2em 0.2em 0.2em,clip=true]{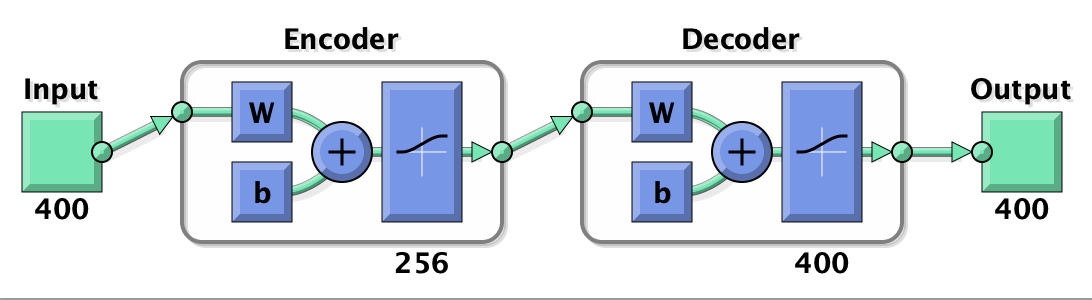}
    \caption{SAE (Layer 3)}
    \end{subfigure}
    \\
    \begin{subfigure}[t]{1\textwidth}\centering
    \includegraphics[width=1\textwidth,trim=1em 0.2em 1em 0.2em,clip=true]{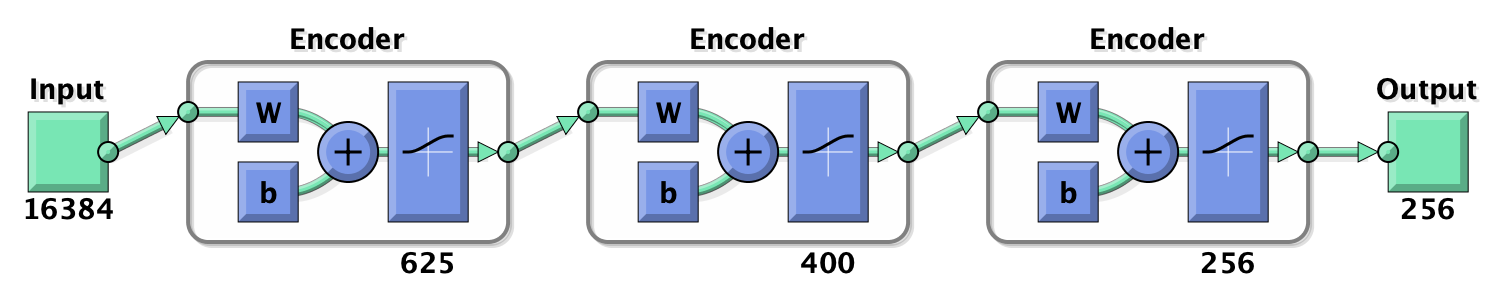}
    \caption{Network of the Stacked Sparse Auto-Encoder (SSAE) used in our strategy.}
    \end{subfigure}
    \caption{Representation of the single layers (first row) for the stacked network (second row) of sparse autoencoders for learning the shape features on profiles in Figure \ref{fig: database 1.0}.}
    \label{fig: stackednet}
    \begin{subfigure}[t]{0.16\textwidth}\centering
    \includegraphics[height=2cm]{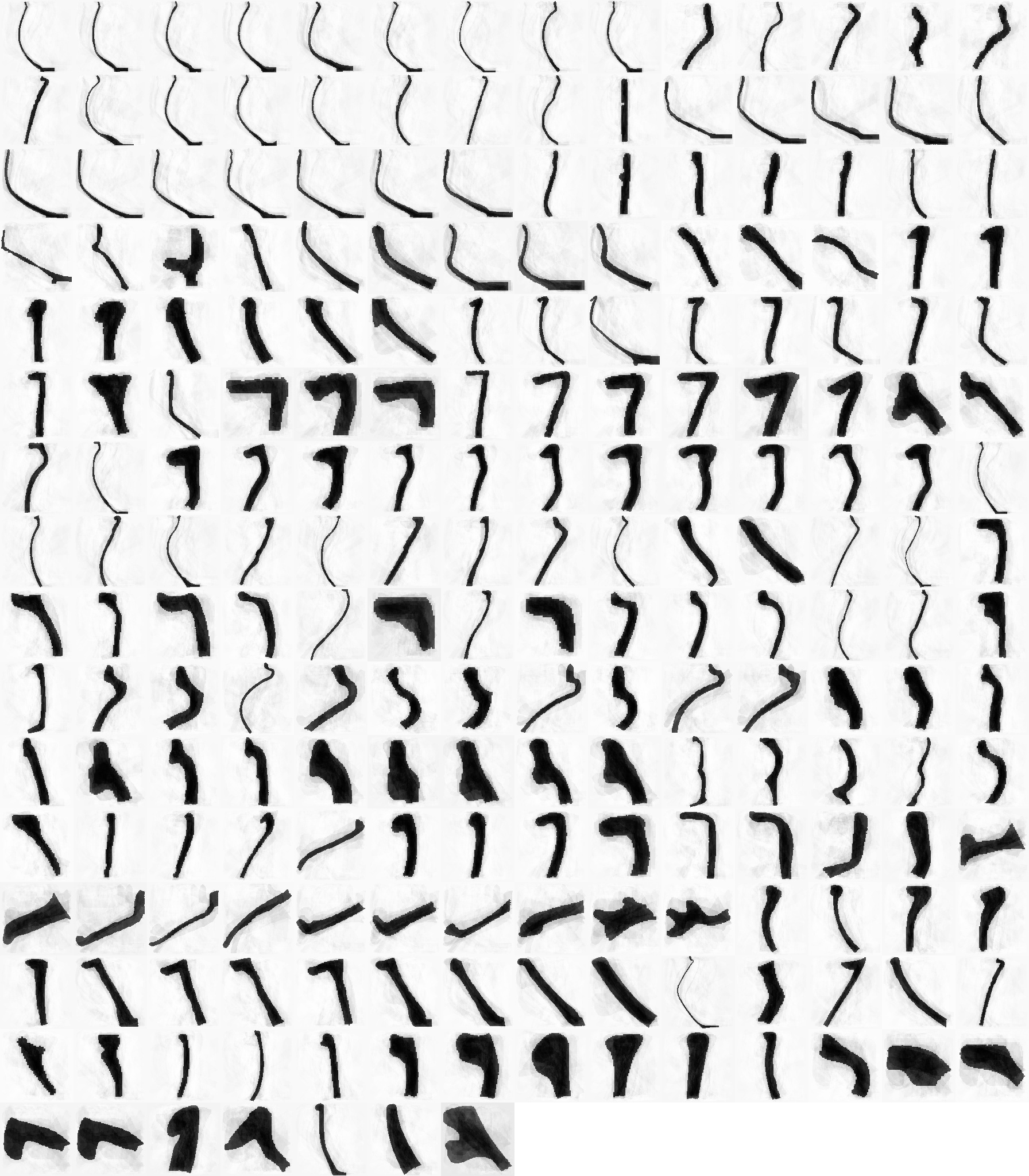}
    \caption{DUN64}
    \end{subfigure}
    \hfill
    \begin{subfigure}[t]{0.16\textwidth}\centering
    \includegraphics[height=2cm]{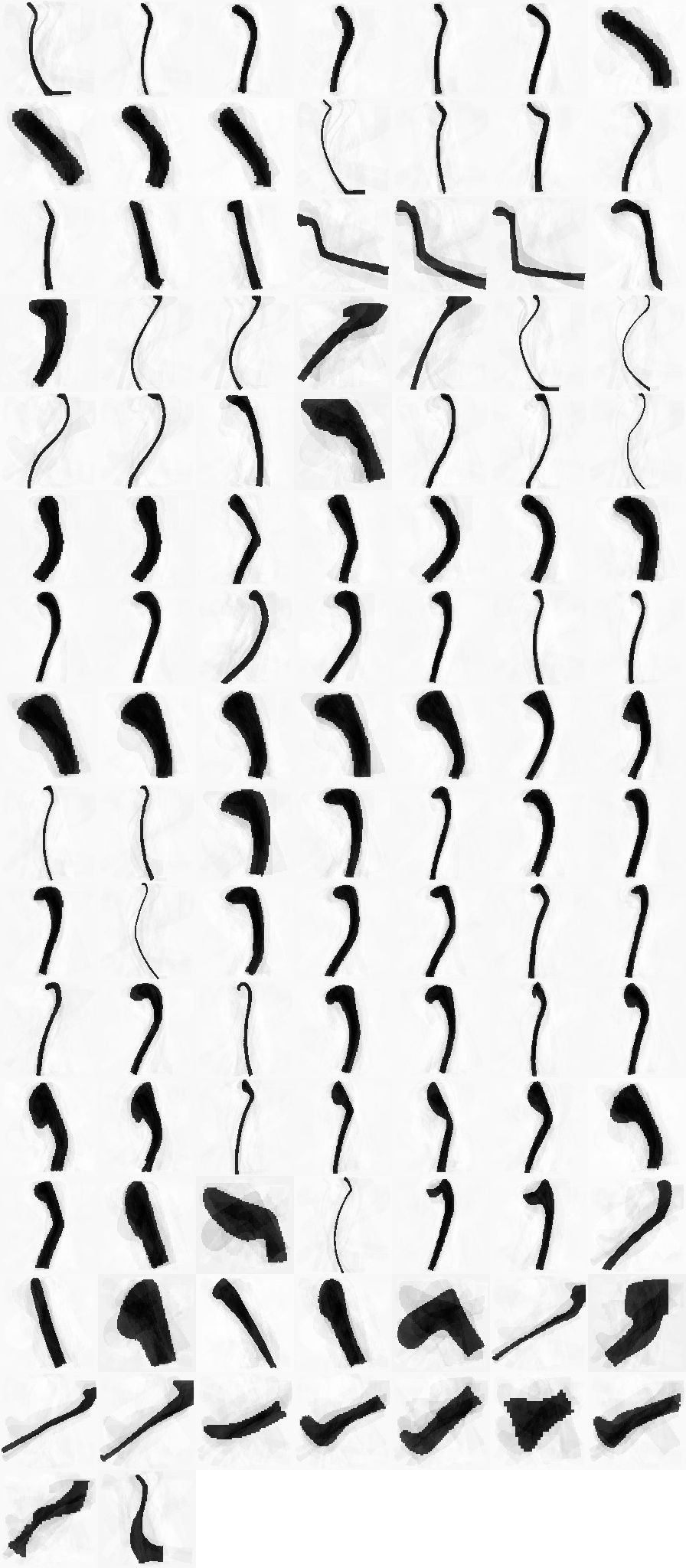}
    \caption{DUN65}
    \end{subfigure}
    \hfill
    \begin{subfigure}[t]{0.16\textwidth}\centering
    \includegraphics[height=2cm]{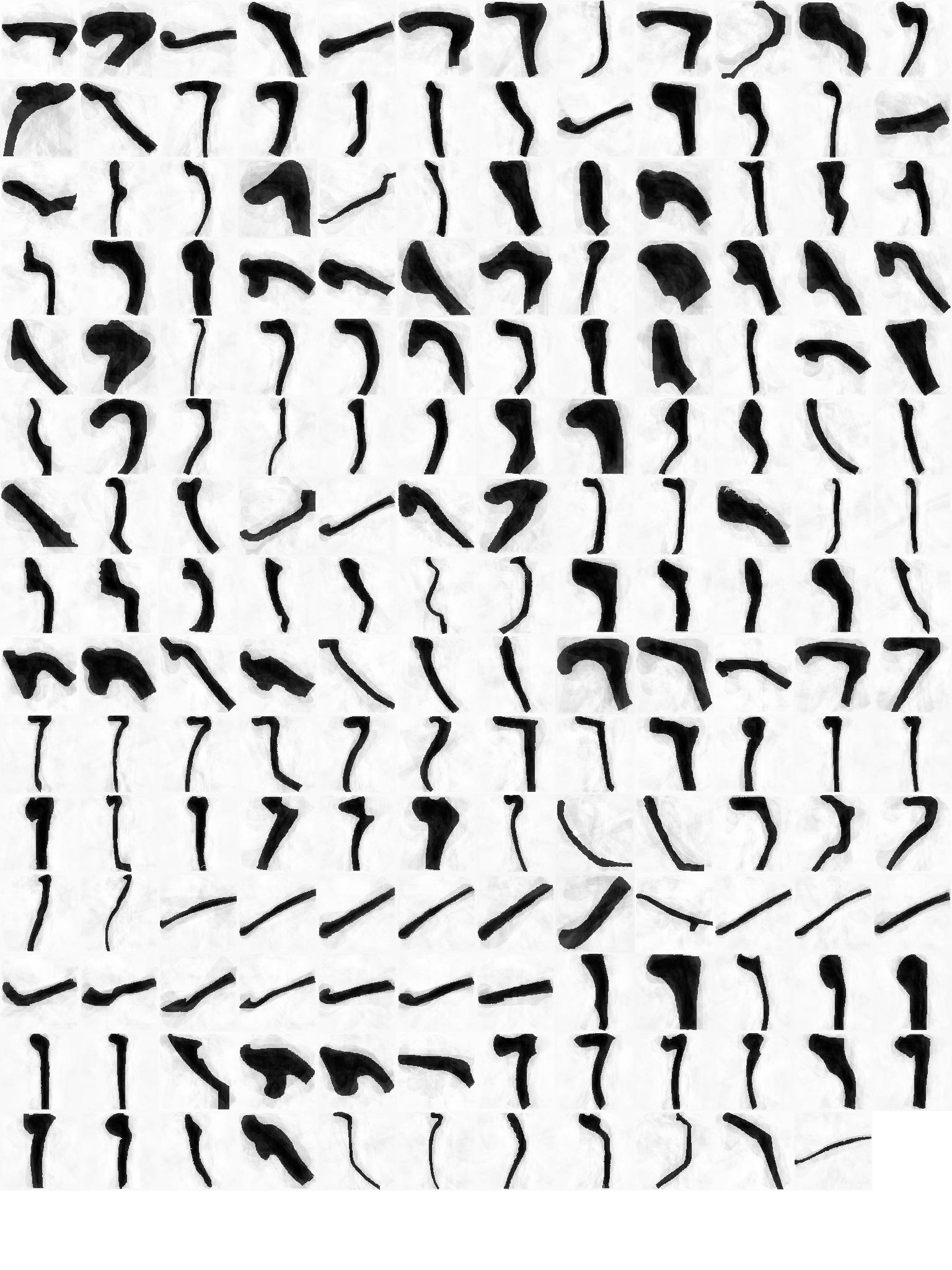}
    \caption{POHL70}
    \end{subfigure}
    \hfill
    \begin{subfigure}[t]{0.48\textwidth}\centering
    \includegraphics[height=2cm]{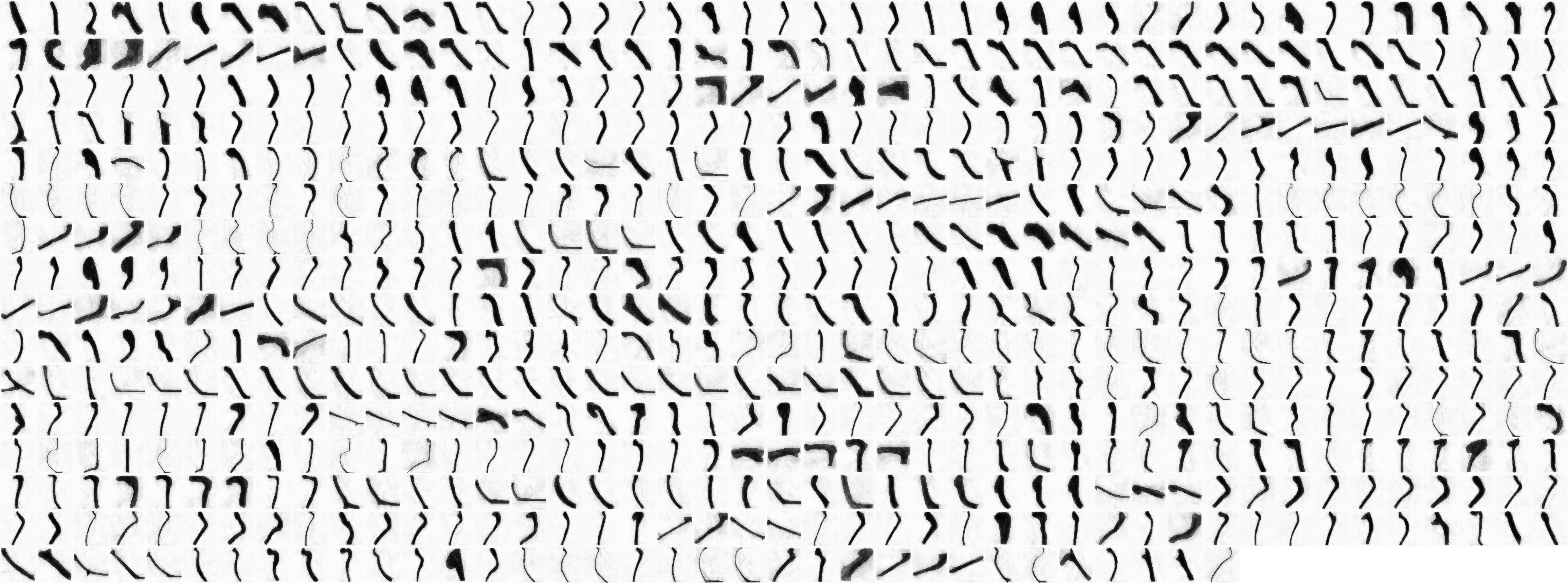}
    \caption{DYS76}
    \end{subfigure}
    \\
    \begin{subfigure}[t]{0.135\textwidth}\centering
    \includegraphics[height=2cm]{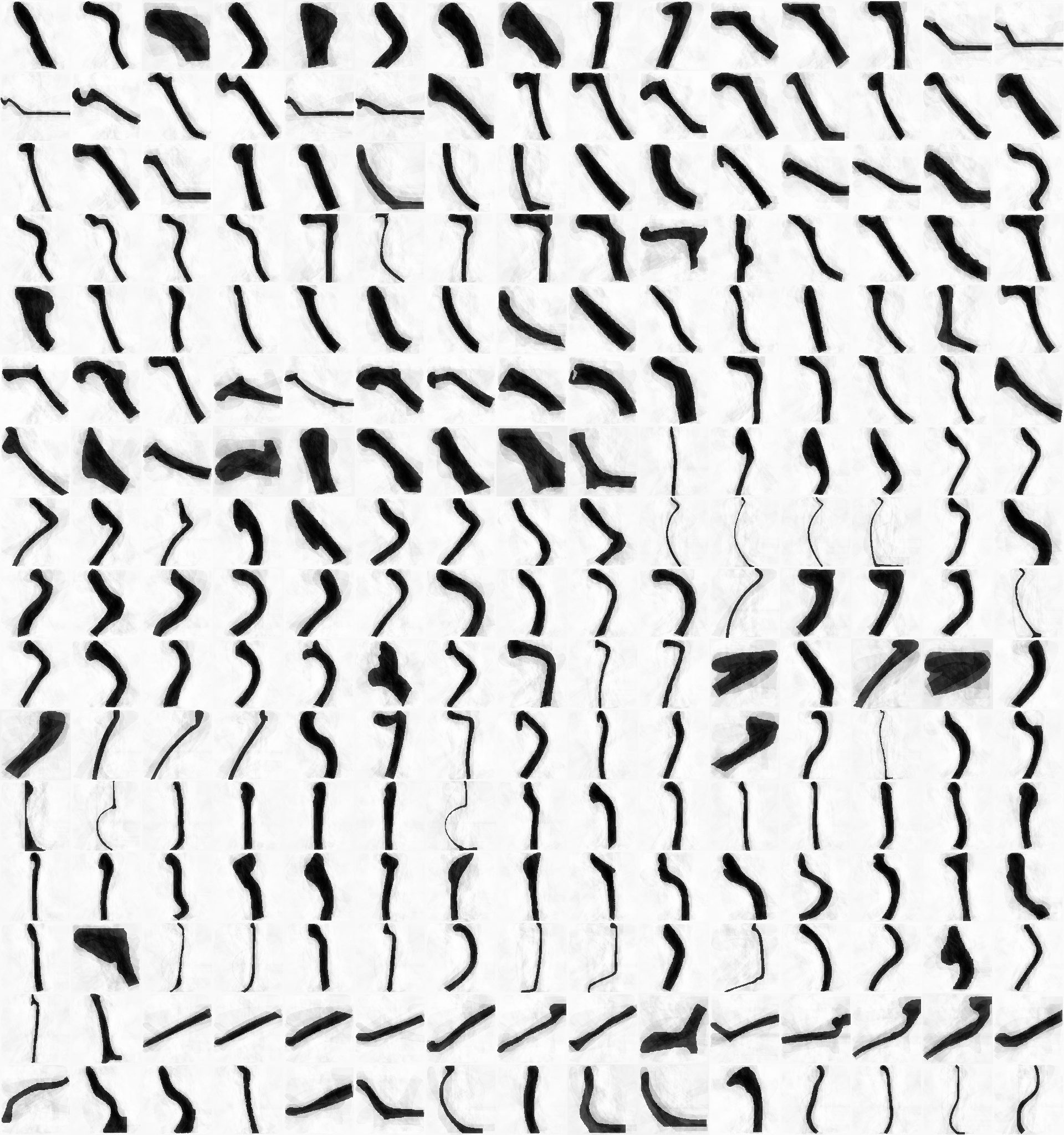}
    \caption{CT84}
    \end{subfigure}
    \begin{subfigure}[t]{0.135\textwidth}\centering
    \includegraphics[height=2cm]{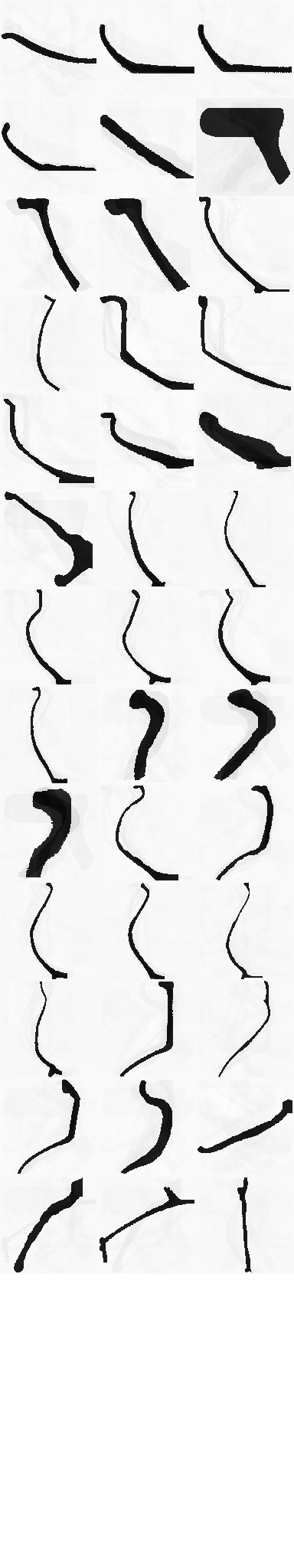}
    \caption{CM91}
    \end{subfigure}
    \begin{subfigure}[t]{0.135\textwidth}\centering
    \includegraphics[height=2cm]{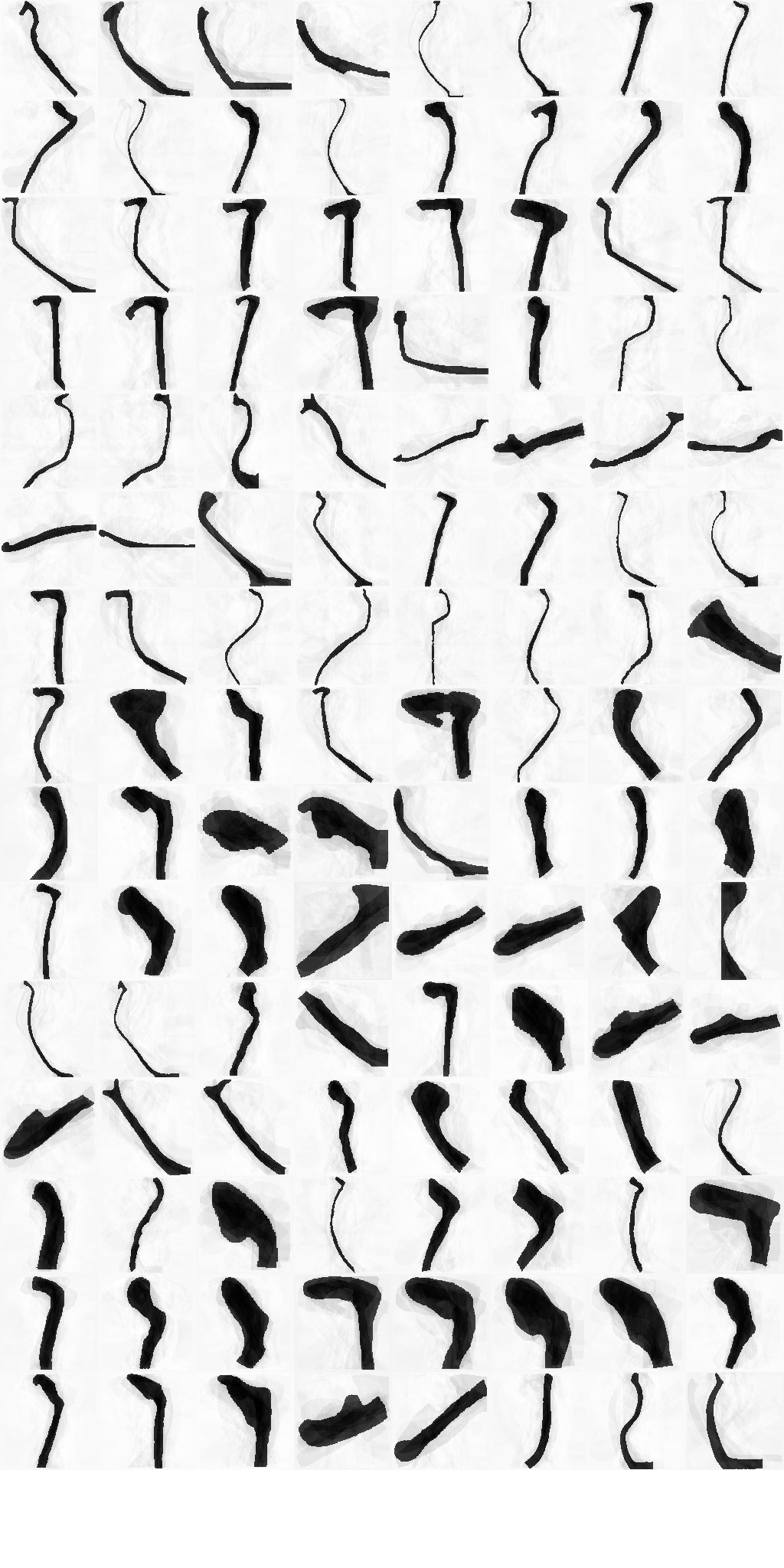}
    \caption{ROB97}
    \end{subfigure}
    \begin{subfigure}[t]{0.135\textwidth}\centering
    \includegraphics[height=2cm]{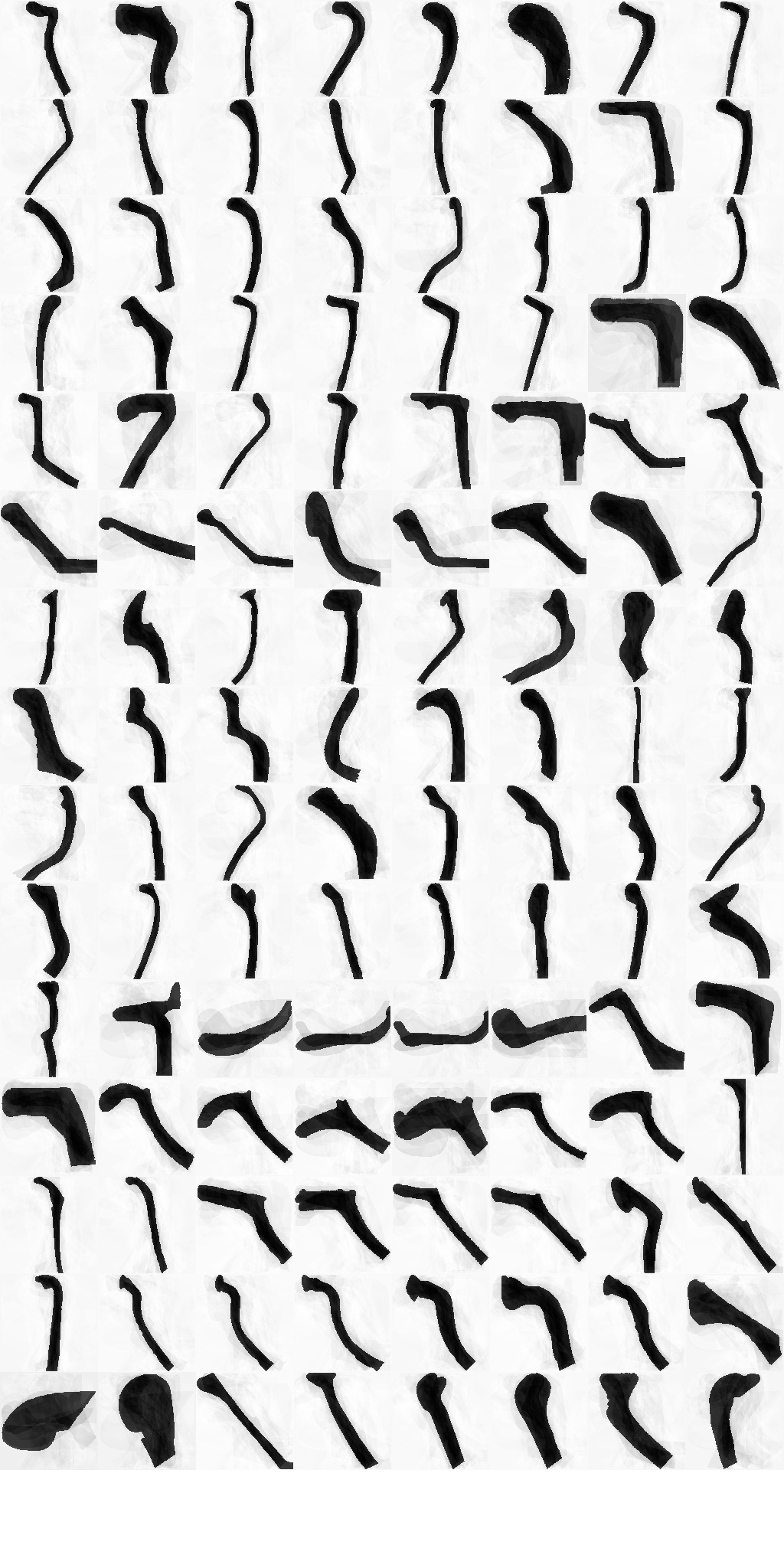}
    \caption{OSTIA1}
    \end{subfigure}
    \begin{subfigure}[t]{0.135\textwidth}\centering
    \includegraphics[height=2cm]{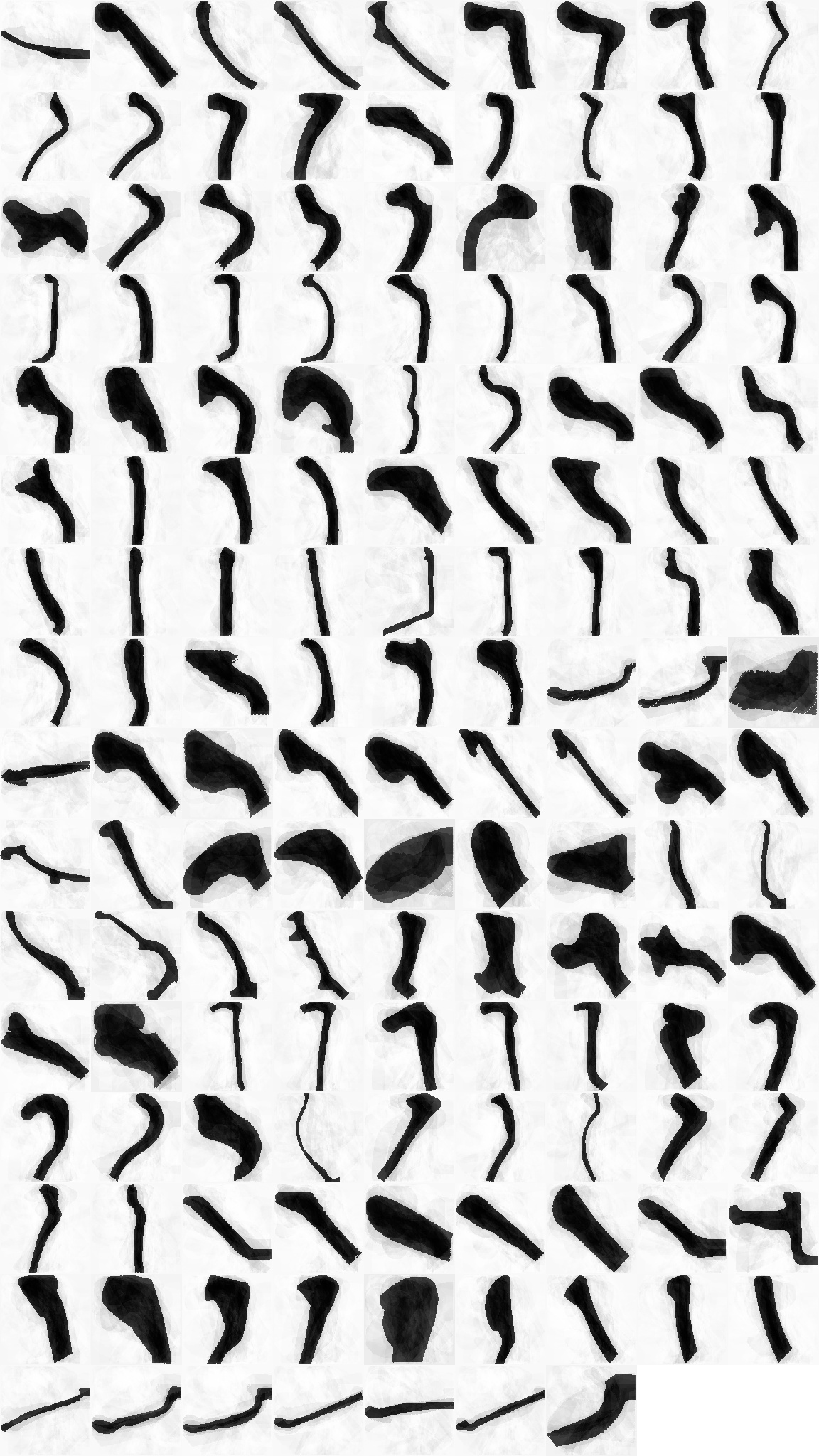}
    \caption{OSTIA2}
    \end{subfigure}
    \begin{subfigure}[t]{0.135\textwidth}\centering
    \includegraphics[height=2cm]{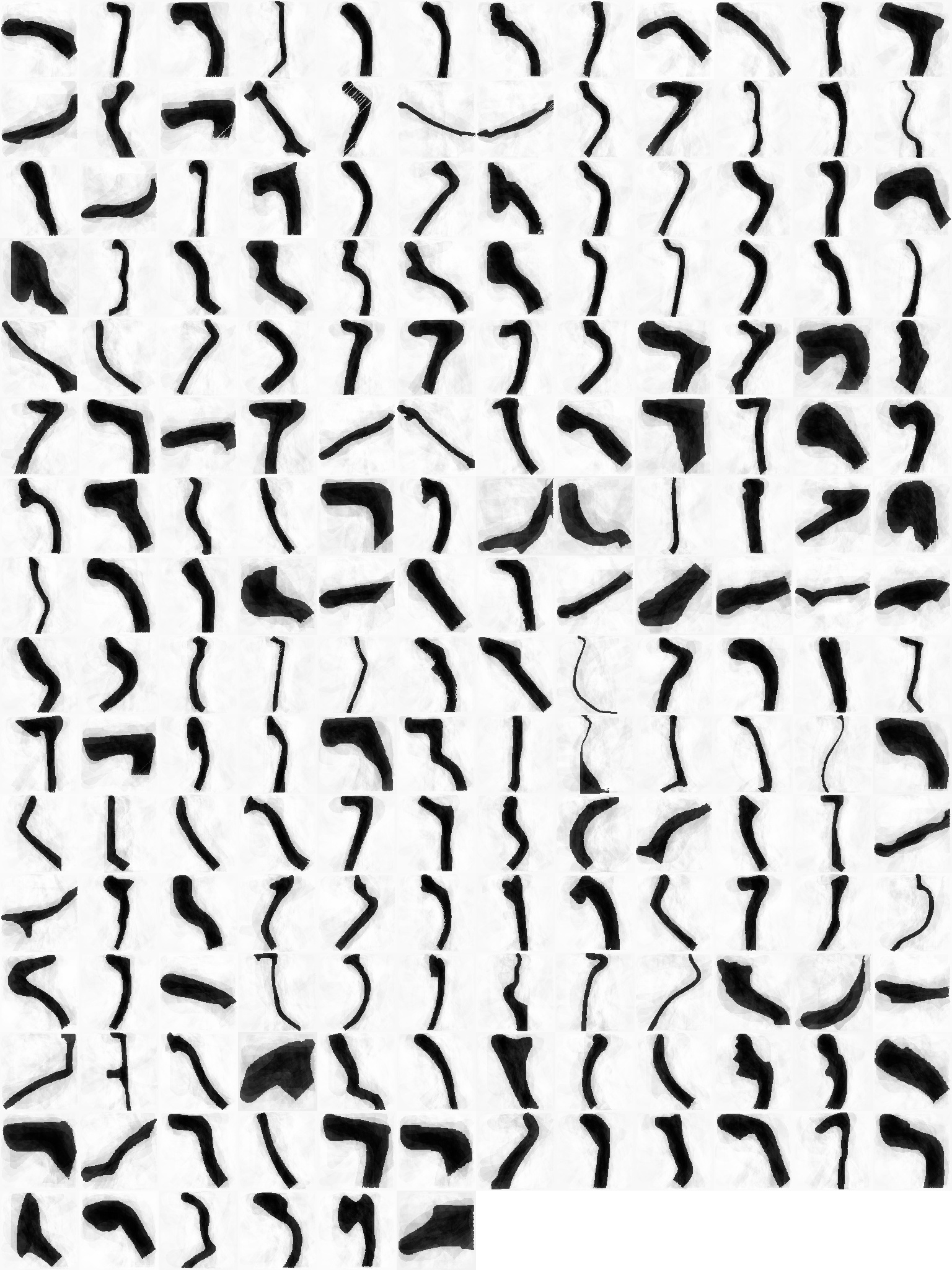}
    \caption{OSTIA3}
    \end{subfigure}
    \begin{subfigure}[t]{0.135\textwidth}\centering
    \includegraphics[height=2cm]{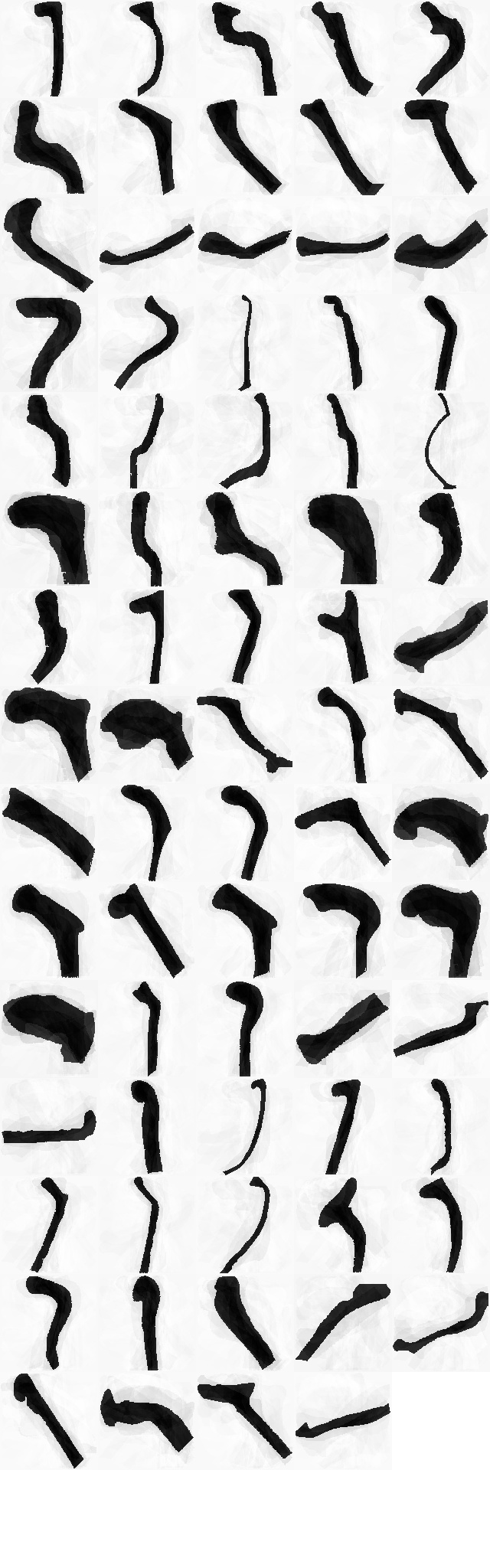}
    \caption{OSTIA4}
    \end{subfigure}
\caption{Reconstruction of Figure \ref{fig: database 1.0} with the SSAE network in Figure \ref{fig: stackednet}.}
    \label{fig: reconstruction database 1.0}
    \centering
    \includegraphics[width=0.75\textwidth,trim=1.5cm 4.25cm 1.5cm 3.5cm,clip=true]{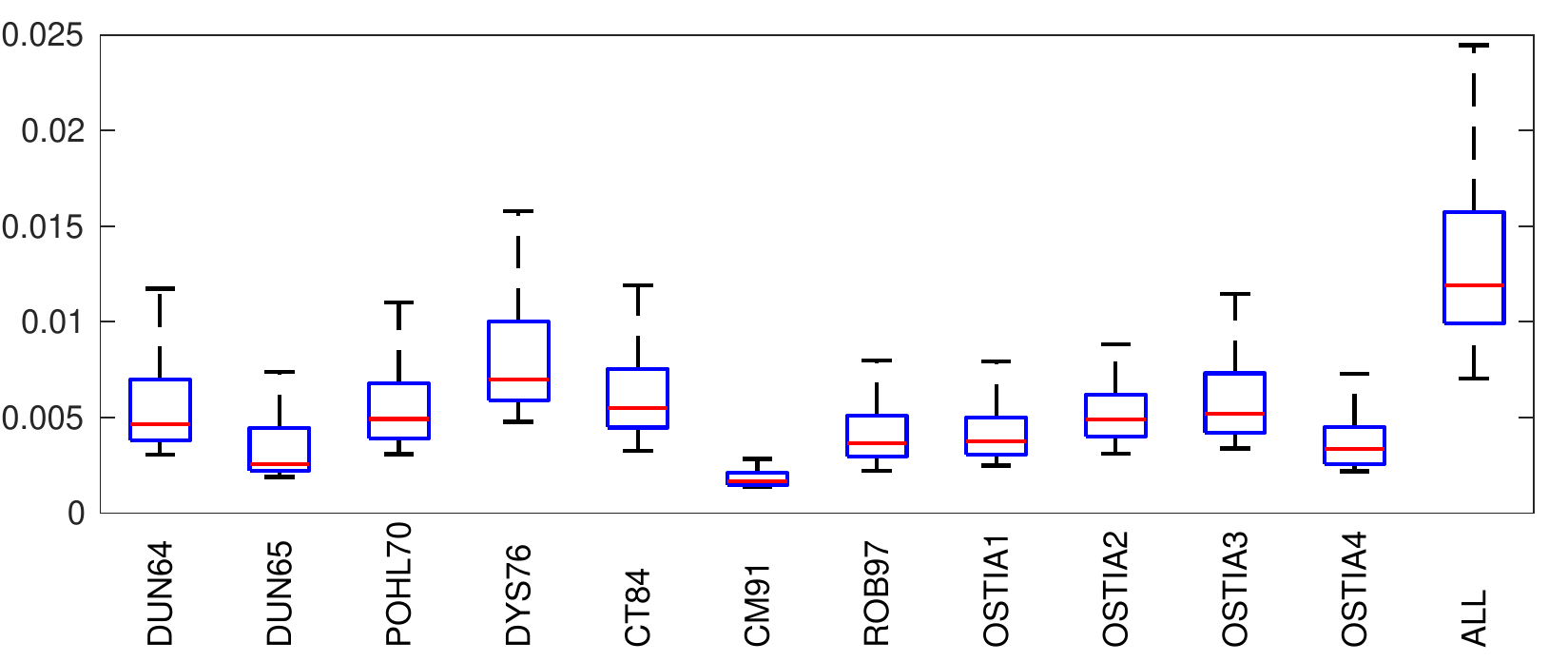}
    \caption{Confidence of the \emph{Mean Squared Reconstruction Error} for each experiment.}
    \label{fig: boxplot}
\end{figure}

Our SSAE network in Figure \ref{fig: stackednet} is composed by 3 stacked layers (each \texttt{IDCAT} experiment is trained with its number of profiles, see Table \ref{tab: database 1.0}) for a maximum number of 500 epochs: the first layer has 625 neurons, the second 400 and the third 256. Each layer learns how to minimise the cost function in \eqref{eq: cost function}, with the \texttt{logsig} transfer function both in the encoder and decoder step and with $\lambda=0.004$ and $\beta=4$ and with 15\% of training examples a neuron reacts to. At the end of the training, we are able to reconstruct in Figure \ref{fig: reconstruction database 1.0} the original profiles of Figure \ref{fig: database 1.0} by means of 256 features, which can now be used in the hierarchical clustering step. We report in Figure \ref{fig: boxplot} the confidence intervals of the \emph{mean squared reconstruction error}: the positive skewness of the boxplots indicates that the frequencies of low errors in the reconstruction occour more often than than large errors.
Both visual and quantitative results confirms that the shapes are reconstructed almost correctly, meaning that the neurons in the latent space are now specialised in the reconstruction of the shapes seen in the database.

\subsubsection{Hierarchical clustering.}
As said, our key idea is to take advantage of the learned features with SSAE for the hierarchical clustering task.
For each experiment, we selected the hierarchical clustering with \texttt{method} and \texttt{metric} among the ones in Table \ref{tab: metric method}, associated to the highest \texttt{cophenet} coefficient: selected choices and values are reported in Table \ref{tab: results single catalogues}, as well as the total number of rims under consideration, the total number of classes obtained and the coherency analysis performed by specialist archaeologists on the so-called \emph{seeds}, i.e.\ the classes with exactly two shapes at the bottom level of the dendrogram.
In most of the cases we observed that best \texttt{cophenet} coefficient is associated to the method \texttt{average} with metric \texttt{chebychev}: such result is not totally unexpected as the \texttt{average} method indicates that all distances between objects from each pair of clusters contribute equally to the mean distance between the clusters while the \texttt{chebychev} metric is based on the maximum coordinate difference between features, effectively useful for a small database of very different rims. 
When the number of data increases, as in the experiments DYS79 and ALL, then the \texttt{cosine} metric is preferred, which is again not surprising since it is a widely used metric in data mining for measuring cohesion within clusters of data \cite{tan2016introduction}.
Final dendrograms for each experiment in Figure \ref{fig: dendrograms}. 

\subsubsection*{Unveiling the invisible profile relations.}
\begin{figure}[htb]
    \centering
    \begin{subfigure}[t]{0.24\textwidth}\centering
    \includegraphics[width=1\textwidth]{\detokenize{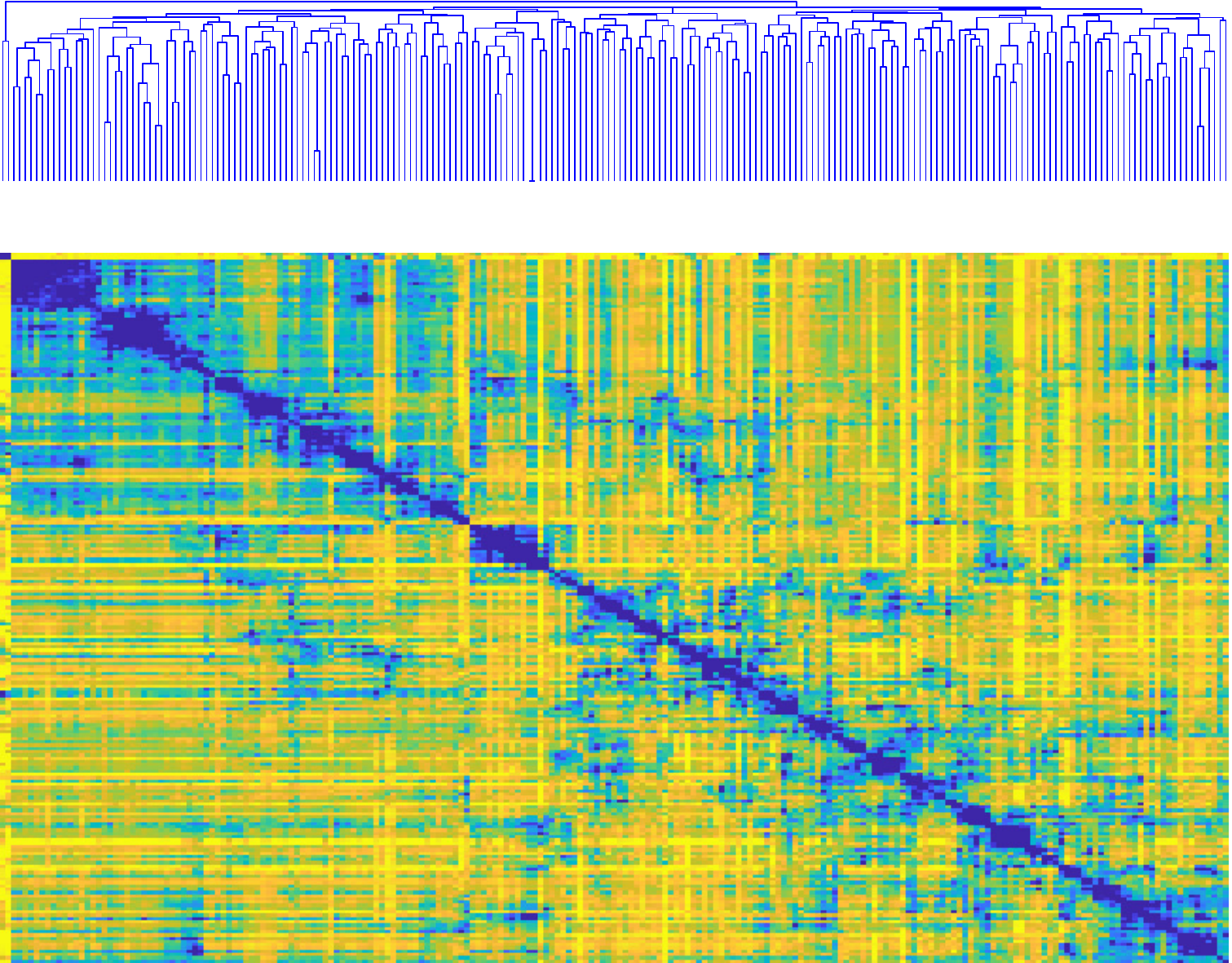}}
    \caption{DUN64}
    \end{subfigure}
    \hfill
    \begin{subfigure}[t]{0.24\textwidth}\centering
    \includegraphics[width=1\textwidth]{\detokenize{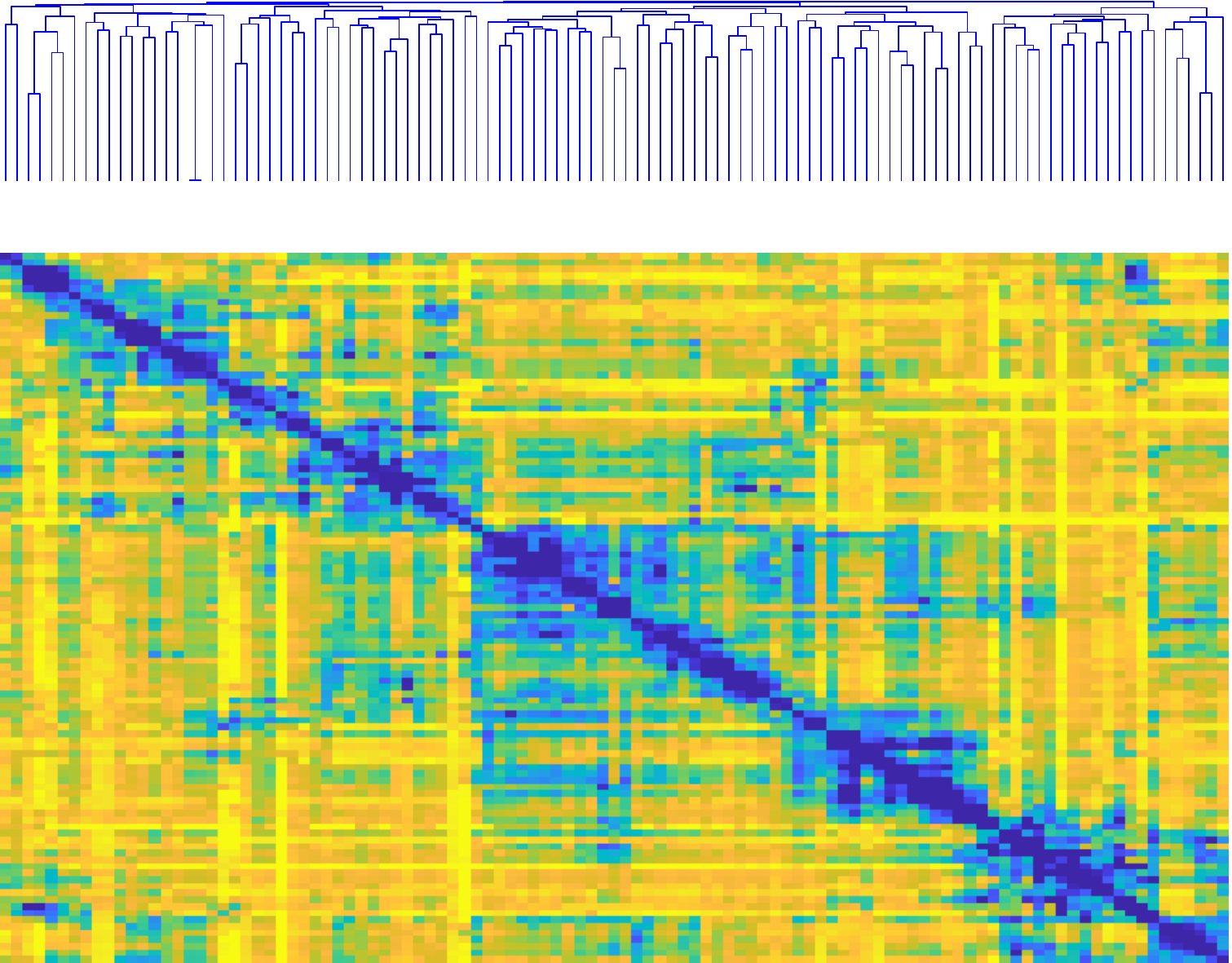}}
    \caption{DUN65}
    \end{subfigure}
    \hfill
    \begin{subfigure}[t]{0.24\textwidth}\centering
    \includegraphics[width=1\textwidth]{\detokenize{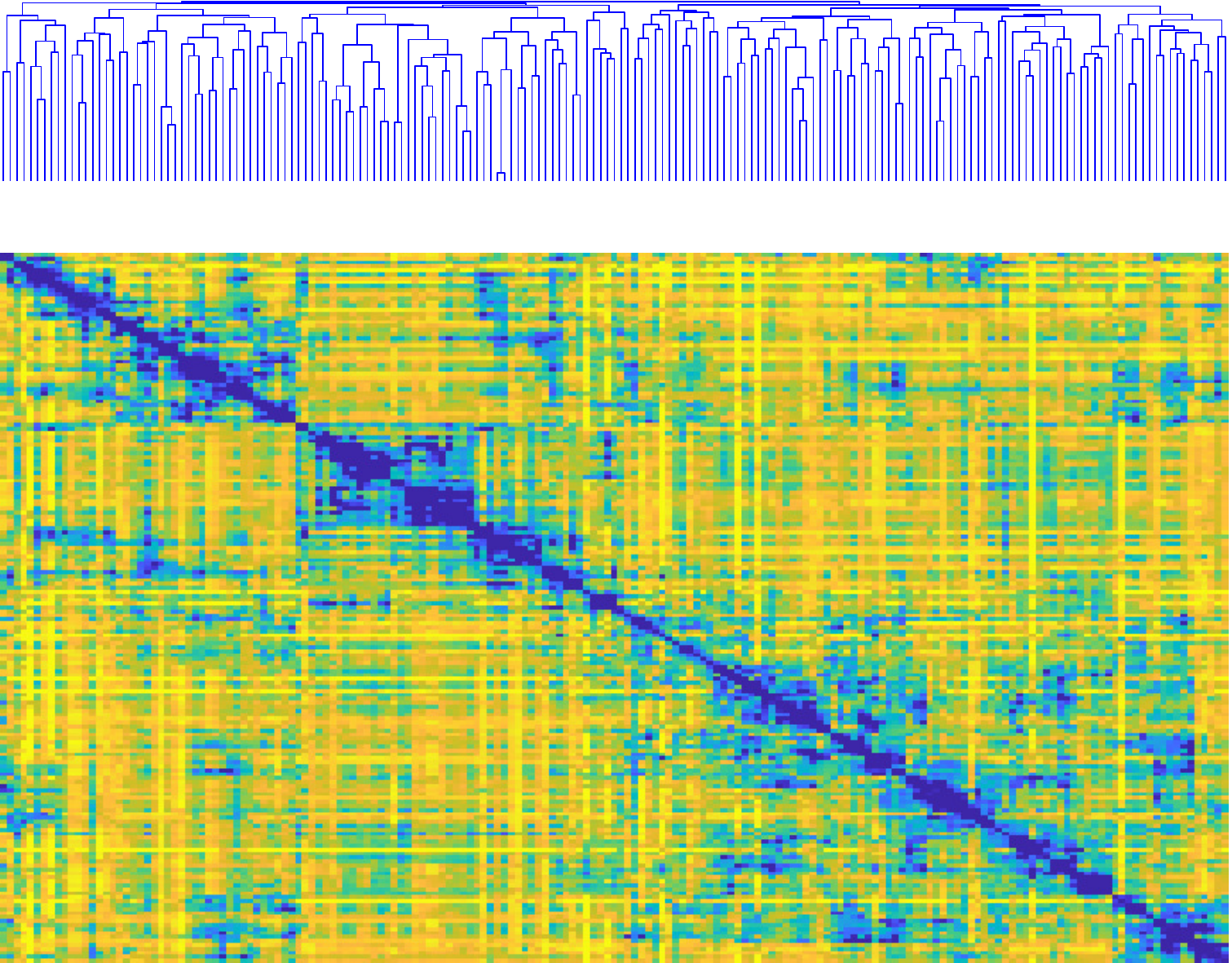}}
    \caption{POHL70}
    \end{subfigure}
    \hfill
    \begin{subfigure}[t]{0.24\textwidth}\centering
    \includegraphics[width=1\textwidth]{\detokenize{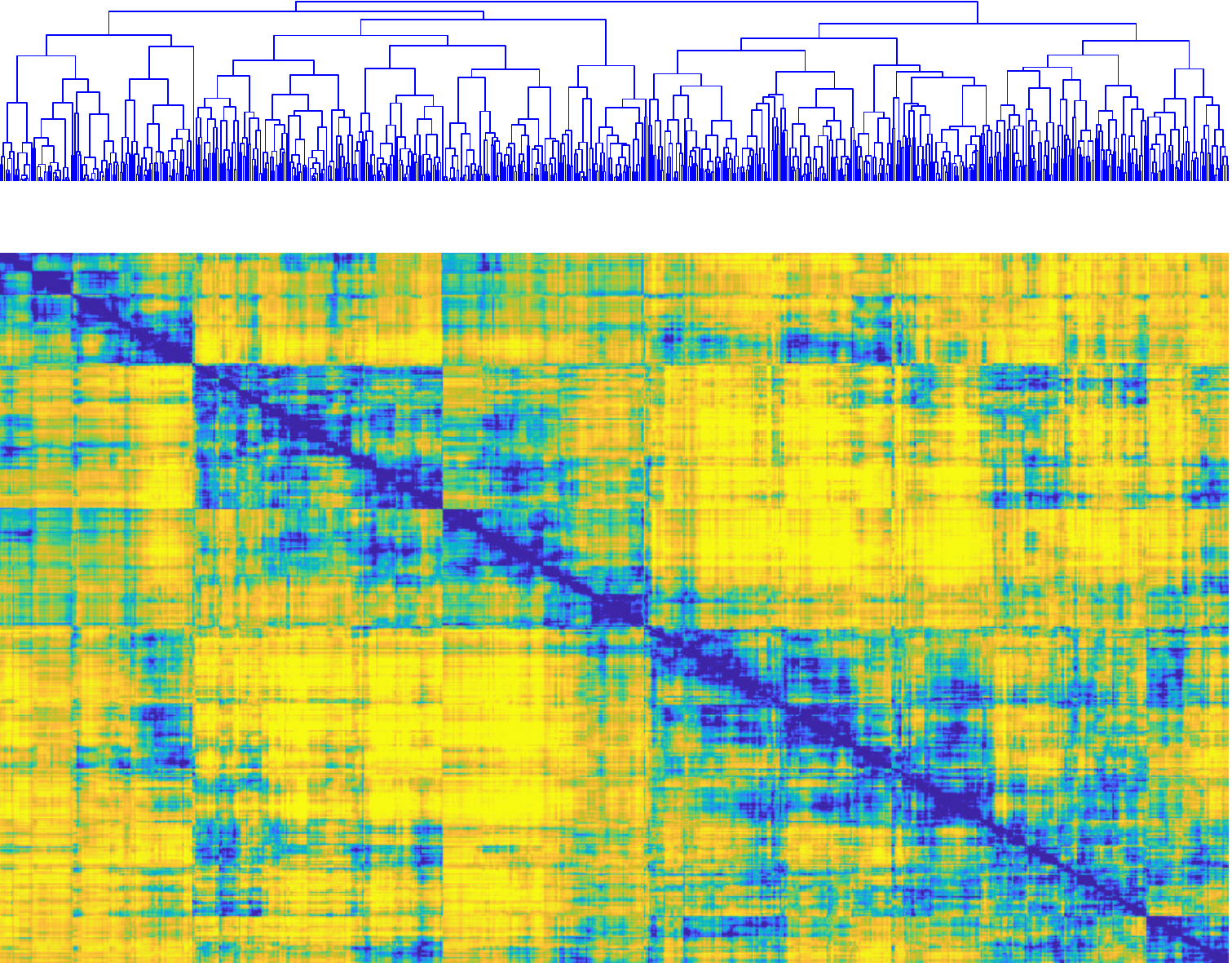}}
    \caption{DYS76}
    \end{subfigure}
    \\
    \begin{subfigure}[t]{0.24\textwidth}\centering
    \includegraphics[width=1\textwidth]{\detokenize{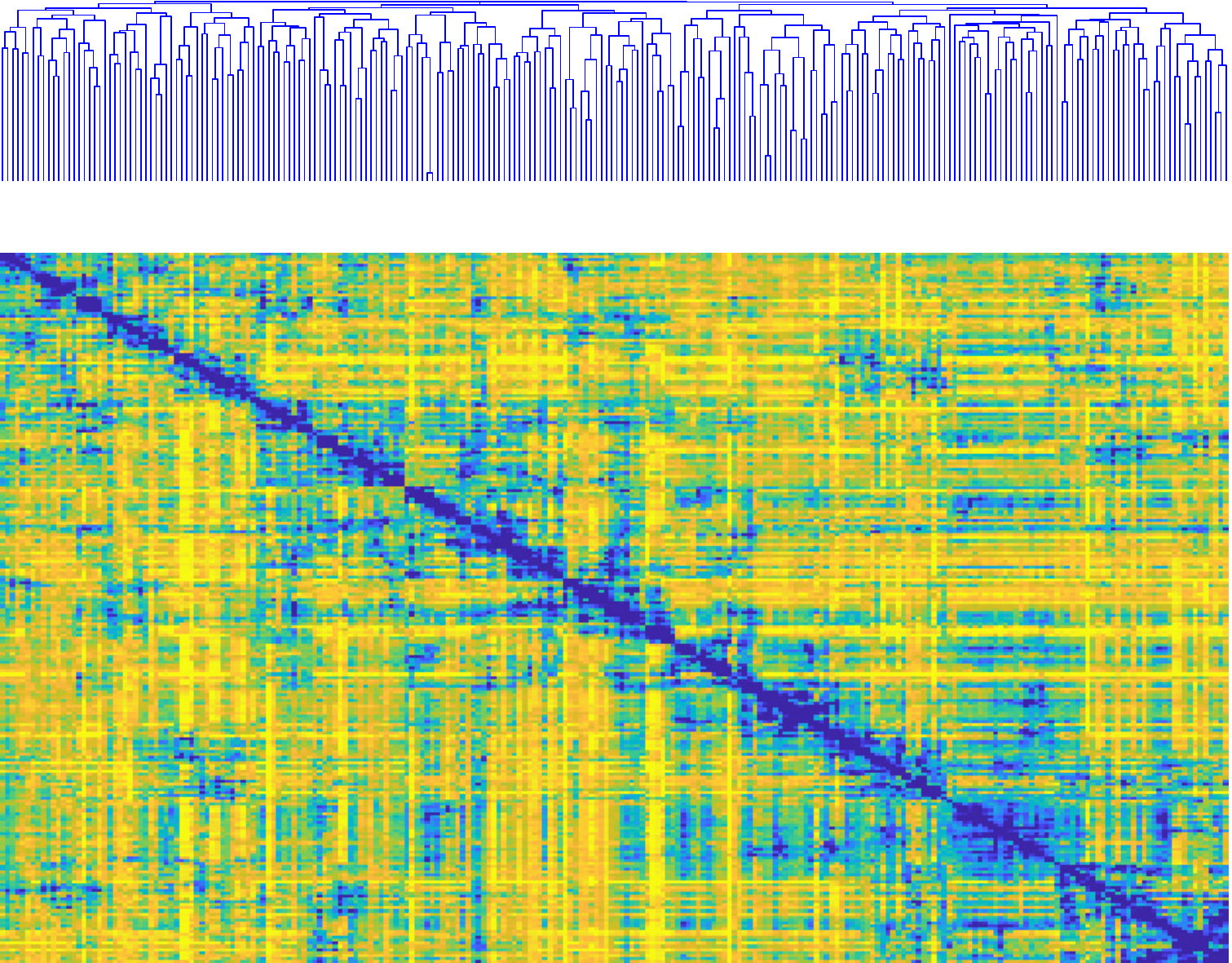}}
    \caption{CT84}
    \end{subfigure}
    \hfill
    \begin{subfigure}[t]{0.24\textwidth}\centering
    \includegraphics[width=1\textwidth]{\detokenize{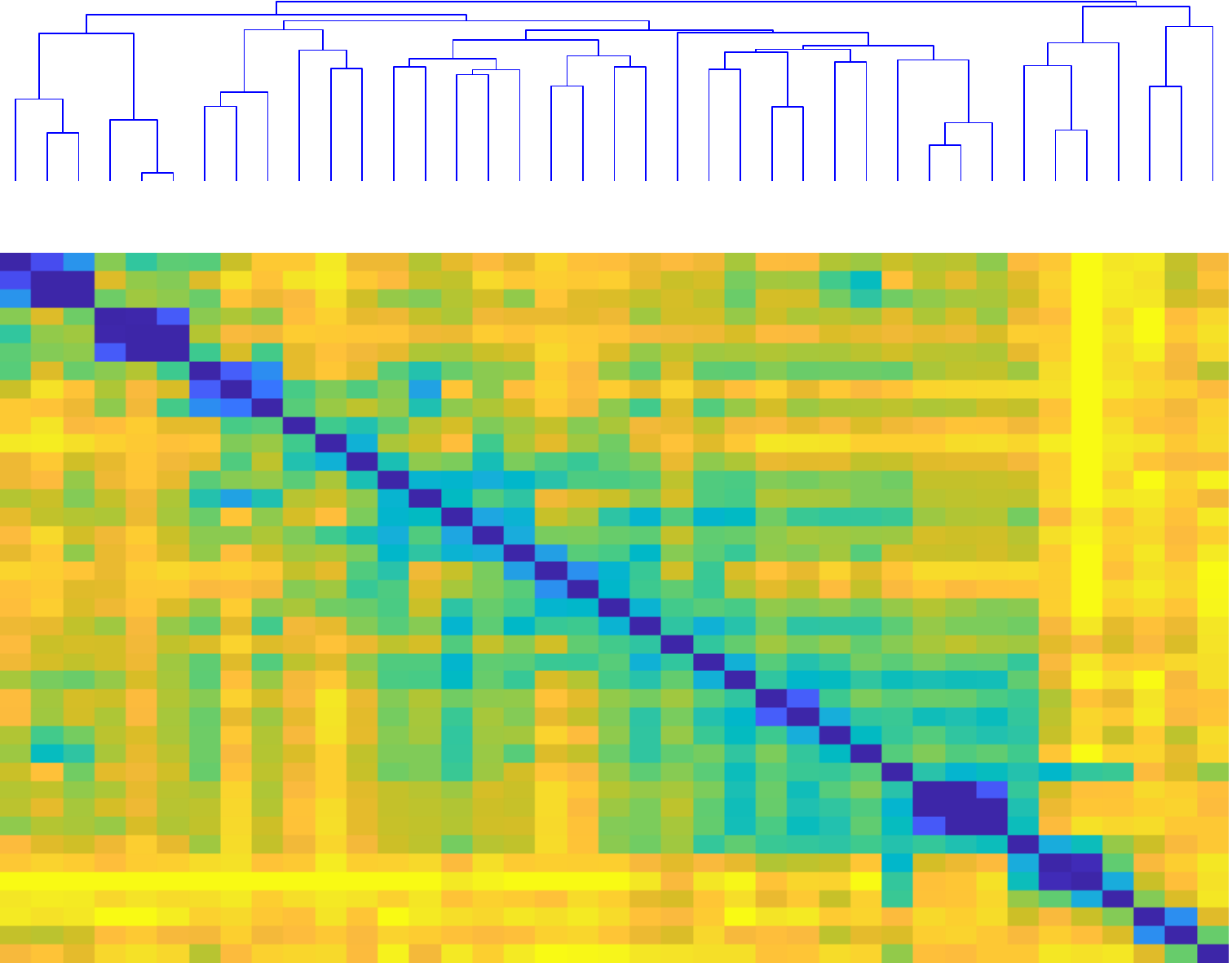}}
    \caption{CM91}
    \end{subfigure}
    \hfill
    \begin{subfigure}[t]{0.24\textwidth}\centering
    \includegraphics[width=1\textwidth]{\detokenize{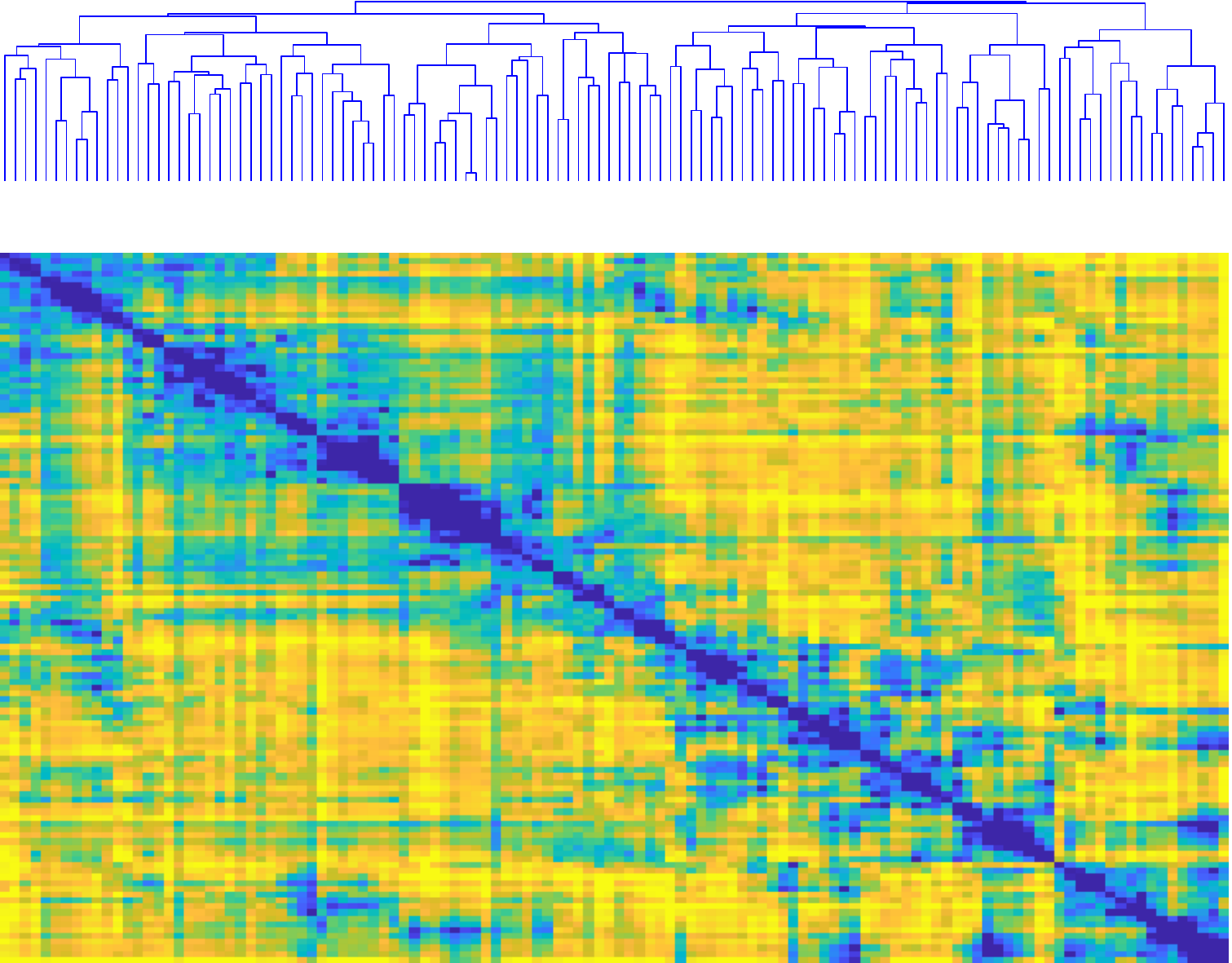}}
    \caption{ROB97}
    \label{fig: dendro ROB97}
    \end{subfigure}
    \hfill
    \begin{subfigure}[t]{0.24\textwidth}\centering
    \includegraphics[width=1\textwidth]{\detokenize{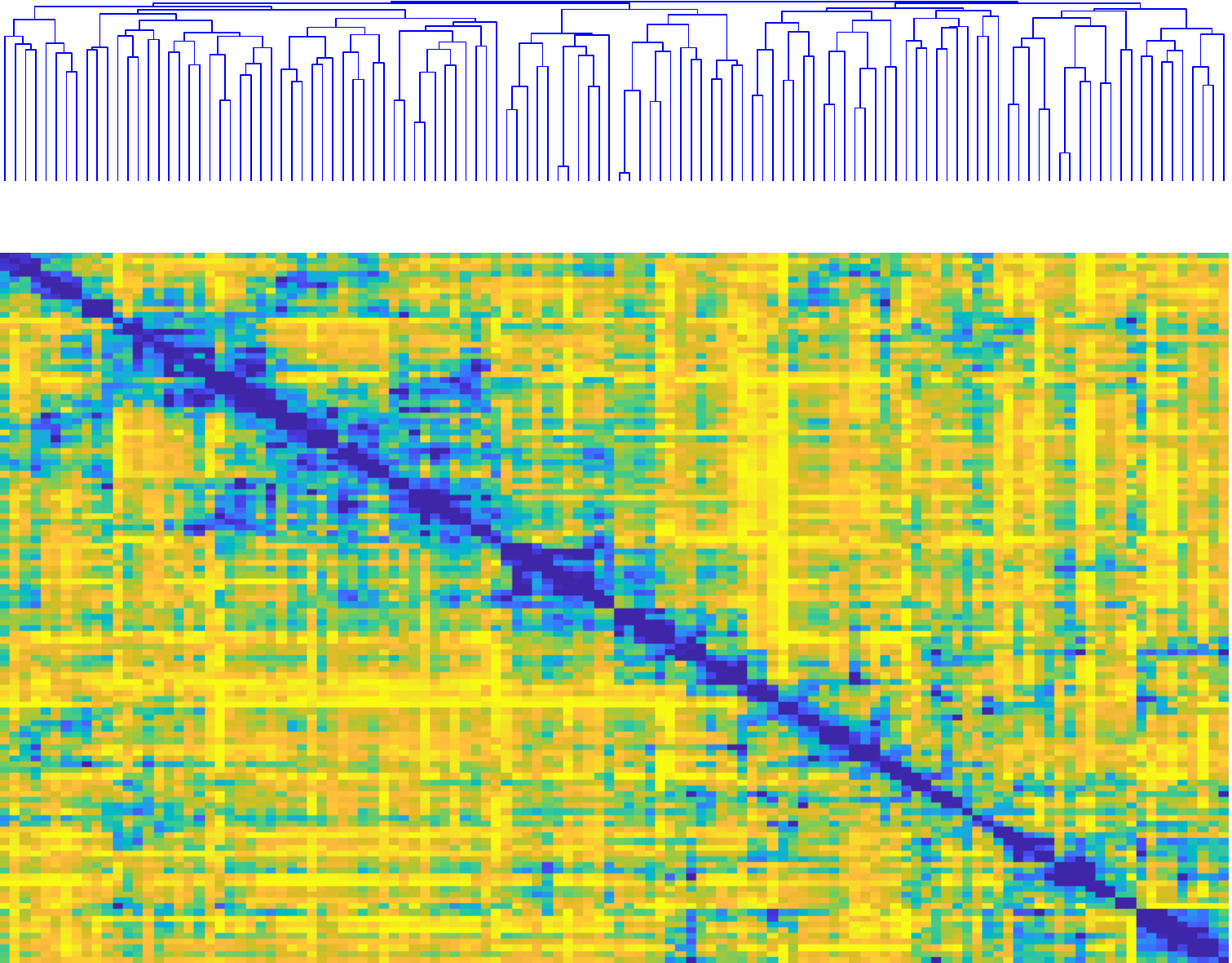}}
    \caption{OSTIA1}
    \end{subfigure}
    \\
    \begin{subfigure}[t]{0.24\textwidth}\centering
    \includegraphics[width=1\textwidth]{\detokenize{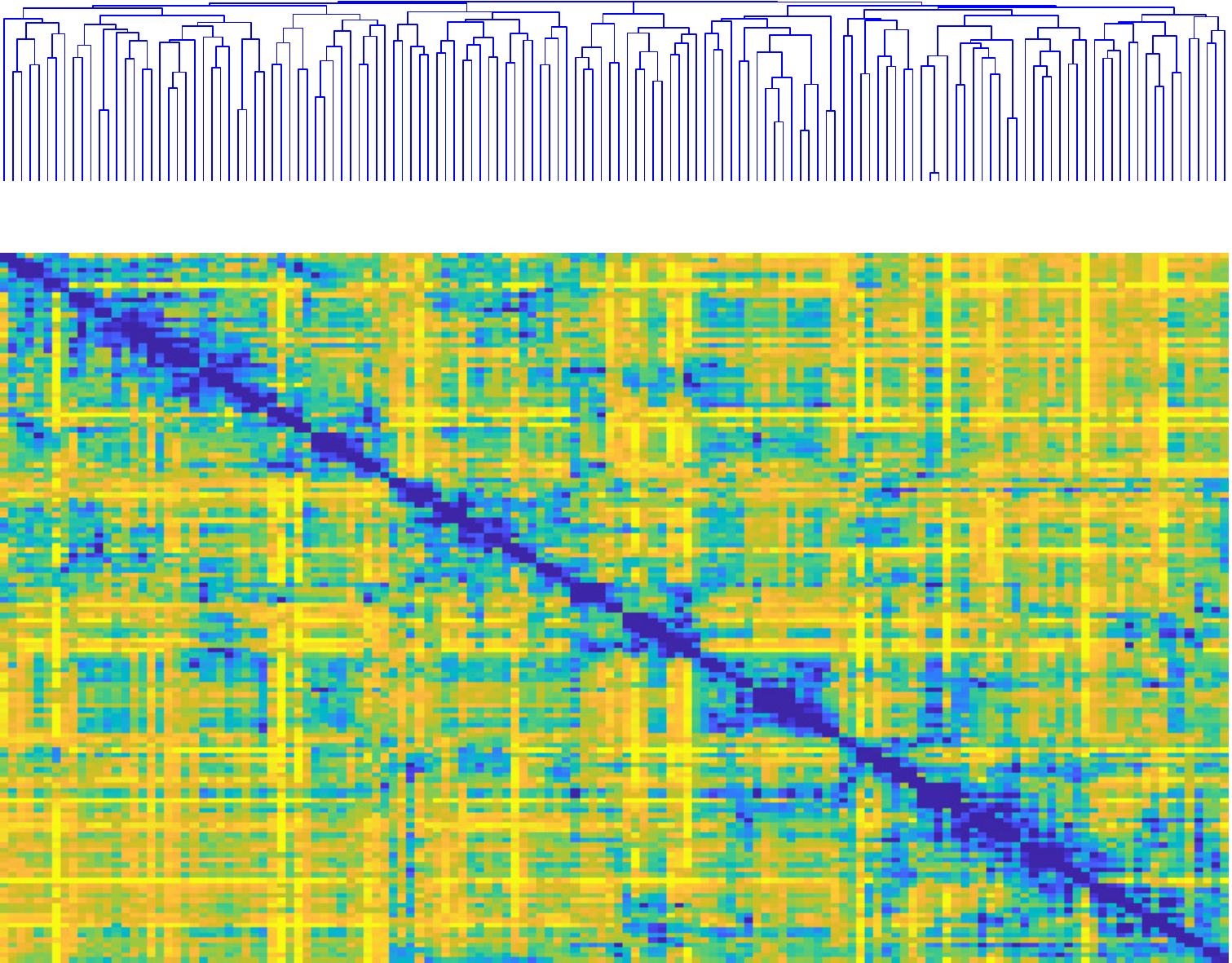}}
    \caption{OSTIA2}
    \end{subfigure}
    \hfill
    \begin{subfigure}[t]{0.24\textwidth}\centering
    \includegraphics[width=1\textwidth]{\detokenize{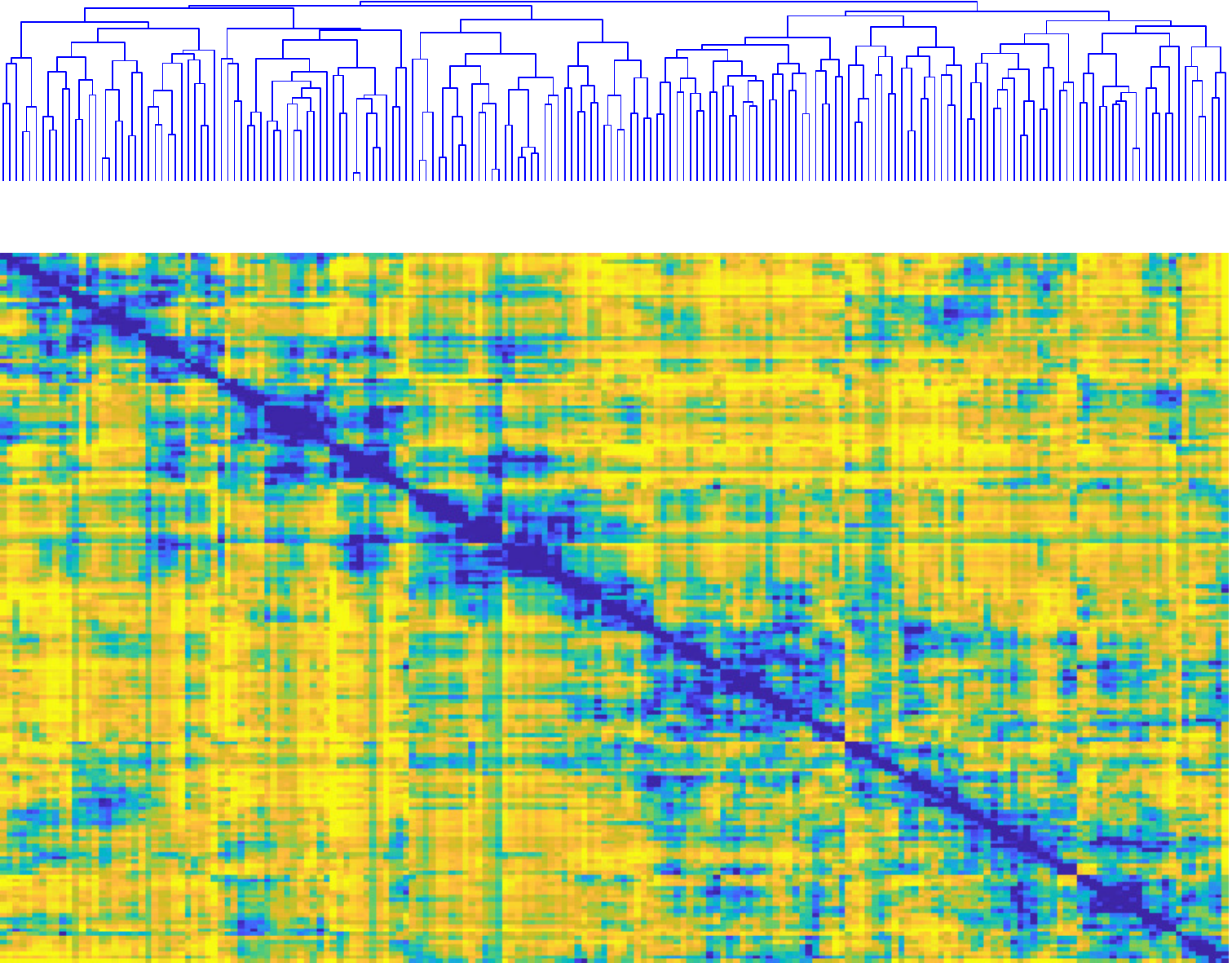}}
    \caption{OSTIA3}
    \end{subfigure}
    \hfill
    \begin{subfigure}[t]{0.24\textwidth}\centering
    \includegraphics[width=1\textwidth]{\detokenize{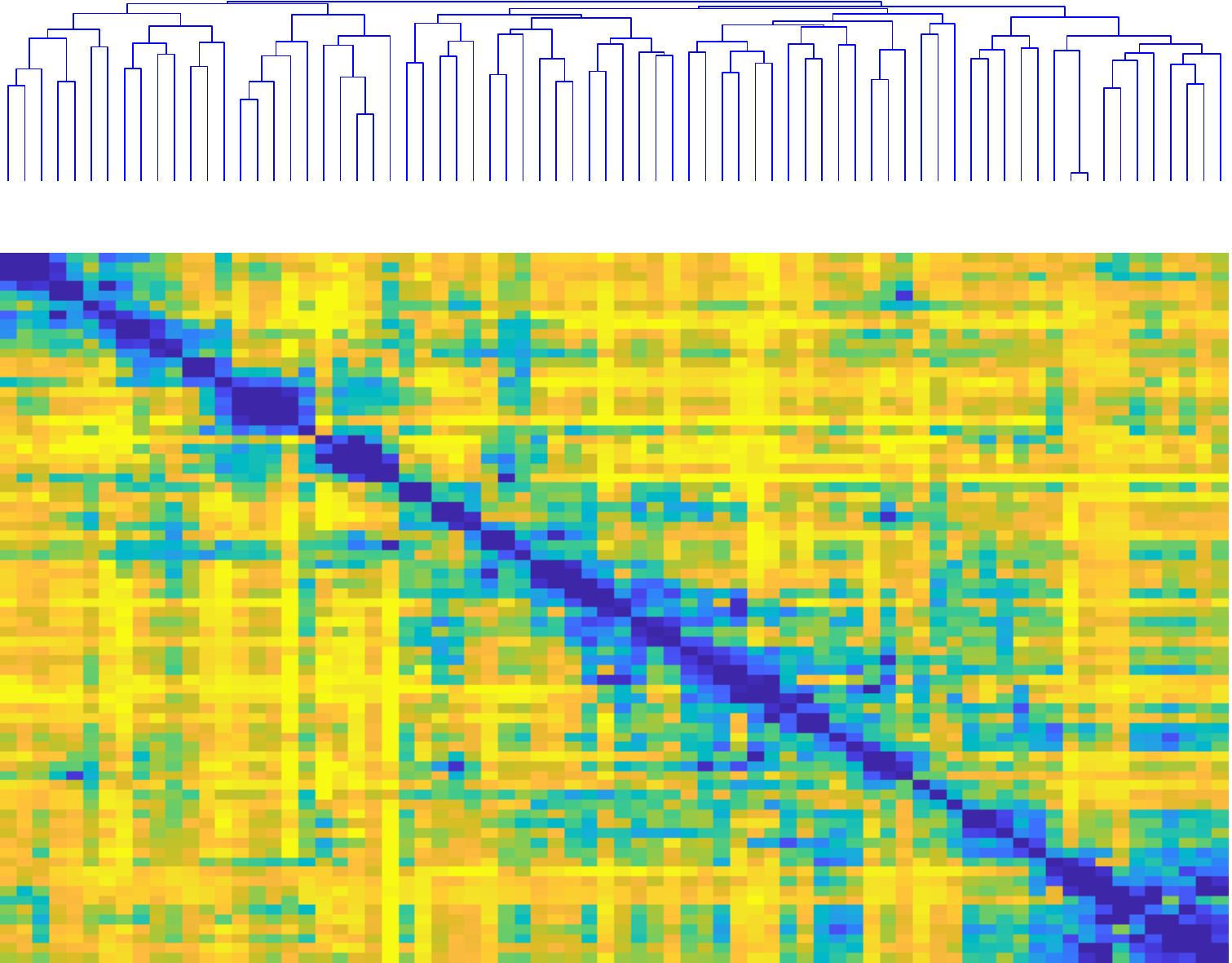}}
    \caption{OSTIA4}
    \end{subfigure}
    \hfill
    \begin{subfigure}[t]{0.24\textwidth}\centering
    \includegraphics[width=1\textwidth,trim=2.6cm 1.6cm 1.9cm 1.2cm,clip=true]{\detokenize{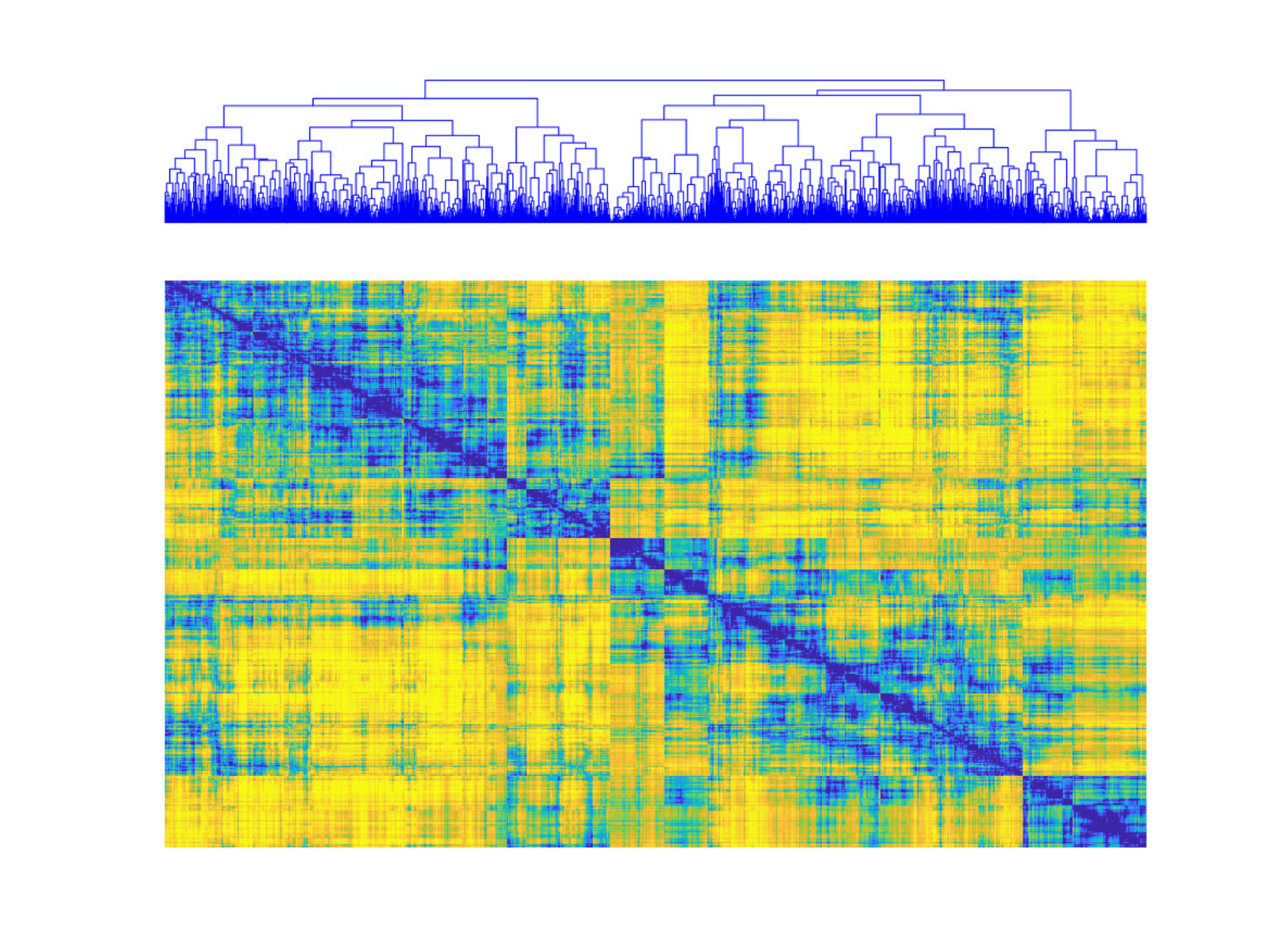}}
    \caption{ALL}
    \label{fig: dendro all}
    \end{subfigure}
    \caption{Dendrograms (top) reflecting the similarity matrix of features (bottom). Colors for similarity matrix are clipped to 3 and 97 percentile of data.}
    \label{fig: dendrograms}
\end{figure}

\begin{figure}[!htb]
    \centering
    \includegraphics[width=0.85\textwidth,trim=6.5cm 1cm 4.75cm 1.5cm,clip=true]{\detokenize{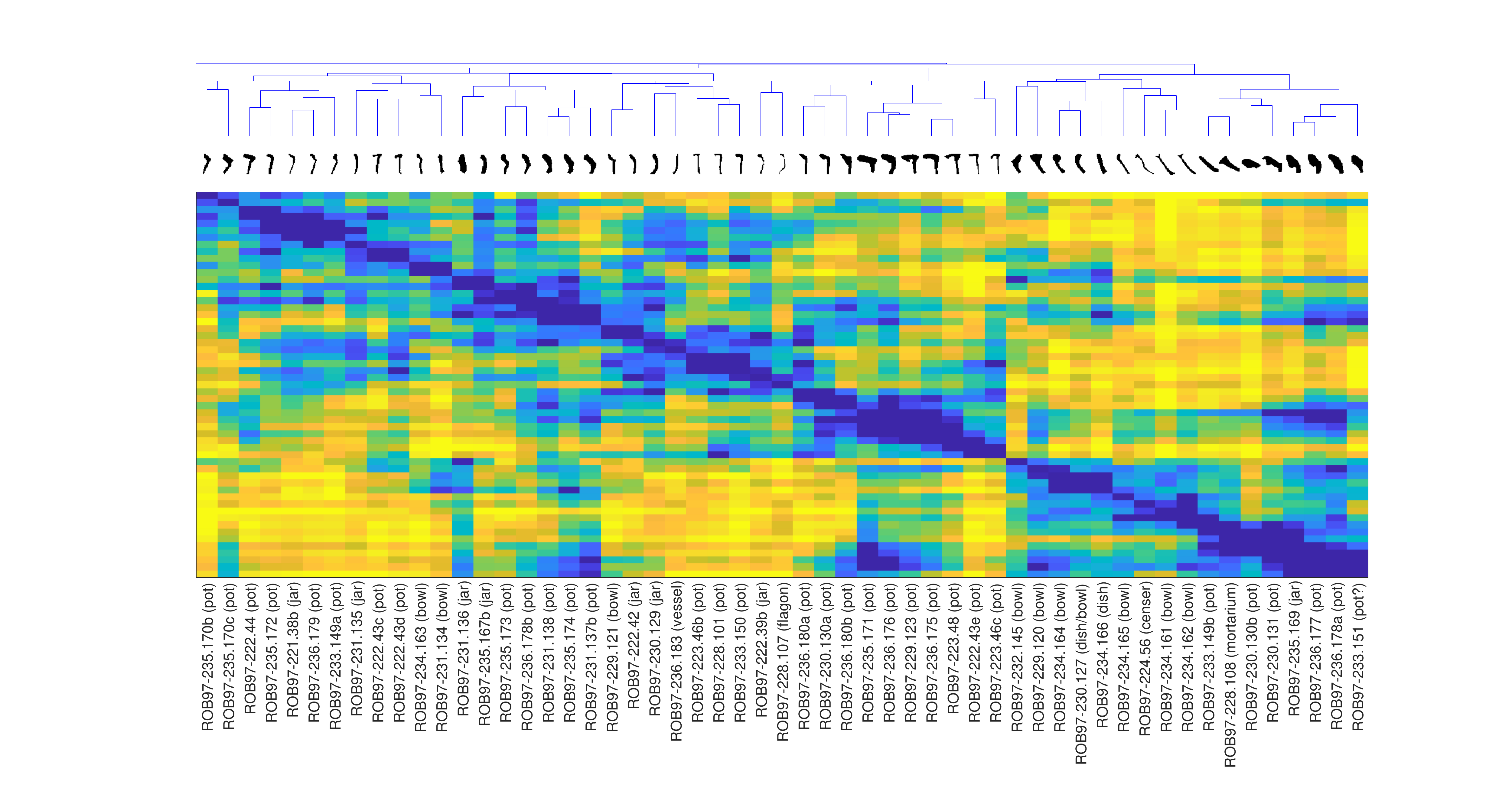}}
    \caption{Zoom on the right-bottom of the  similarity matrix in Fig.\ \ref{fig: dendro ROB97} (ROB97), associated dendrogram and shapes: similarity is unveiled also by metadata, e.g.\ proximity of types (like pots, jars and bowls) and string-name of profiles.}
    \label{fig: dendro-zoom 1}
    \includegraphics[width=0.85\textwidth,trim=6.5cm 1cm 4.75cm 1.5cm,clip=true]{\detokenize{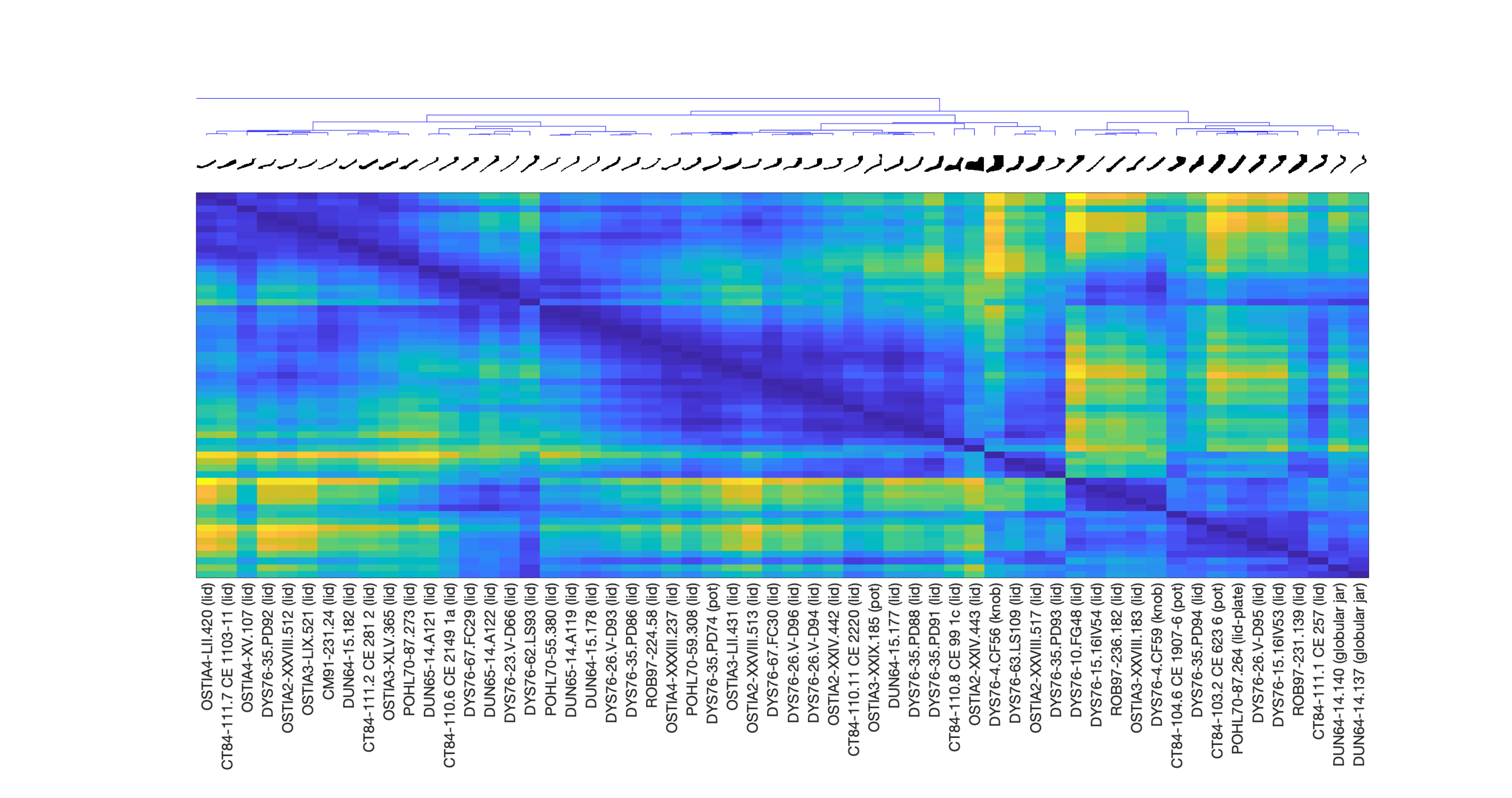}}
    \caption{Zoom on the right-bottom of the similarity matrix in Fig.\ \ref{fig: dendro all} (ALL), associated dendrogram and shapes: almost all profiles are recognised as lids from different corpora.}
    \label{fig: dendro-zoom 2}
\end{figure}

Given the cross-discipline nature of the collaboration described in this paper, we aim to explore possible shared research directions for data scientists and archaeologists. 
On top of the already established classification provided by the authors of each catalogue, we may be prompted to unveil new similarities between profiles, so as to hand in to experts an additional tool for the inspection of pottery catalogues. 
However, the lack of a commonly accepted “ground truth" in this type of data and their partiality has an impact on the final evaluation of the performances. 
Nevertheless, from a data scientist point of view, we can see in Figure \ref{fig: dendrograms} that the similarity matrix of each experiment very well reflects the dendrogram structure, where tiny blocks on the matrix diagonal appears due to comparable features. This is particularly clear on catalogues with less data, e.g.\ CM91 or ROB97, while some macro-blocks, i.e.\ subclusters of many shapes, are particularly evident in DYS76. 
These plots also reflects the fact of the high variance in a small amount of data and therefore only the so-called ``seeds'' (with few expansion steps towards the top of the dendrogram) can provide a reliable cluster of similar shapes as checked in the results summarised in Table \ref{tab: results single catalogues}.
The pairing of seeds appears very promising in so far as the workflow was able to bring together similar profiles among coherent sub-datasets (e.g.\ dish-lids), see e.g.\ the zoom of dendrograms with pottery profiles in Figures \ref{fig: dendro-zoom 1} and \ref{fig: dendro-zoom 2}. However, the inclusion of profiles derived from both intact and partial vessels introduces an element of confusion which the system could not effectively deal with yet (e.g.\ it compares a rim-fragment with an entire pot). 
For the effectiveness of our approach in supporting the classification of the database, it will be necessary to identify vessel parts (rim, wall, base) in each profile and introduce this level of information in the proposed dataset \texttt{ROCOPOT}: this is left for future work.

\vspace{-2em}
\begin{table}[!htb]
\caption{Details of results on Rims.}\scriptsize
\label{tab: results single catalogues}
\begin{tabularx}{1\textwidth}{Xccccc|cc|cc}
\\
\toprule
\multirow{2}{*}{\texttt{IDCAT}} & \multirow{2}{*}{Rims} & \multirow{2}{*}{\texttt{method}} & \multirow{2}{*}{\texttt{metric}} & \multirow{2}{*}{\texttt{cophenet}} & \multirow{2}{*}{Classes} & \multicolumn{4}{c}{Coherency: \#/Seeds (\%)}\\
& & & & & & \multicolumn{2}{c}{Positive} & \multicolumn{2}{c}{Negative} \\
\midrule
DUN64   &  217 & \texttt{average} & \texttt{chebychev}  & 0.7217 &  216 &  61/79 & (77\%) & 18/79 & (13\%) \\
DUN65   &  107 & \texttt{average} & \texttt{chebychev}  & 0.7699 &  106 &  30/37 & (81\%) & 07/37 & (19\%) \\
POHL70  &  179 & \texttt{average} & \texttt{chebychev}  & 0.7081 &  178 &  35/65 & (54\%) & 30/65 & (46\%) \\
DYS76   &  679 & \texttt{average} & \texttt{cosine}     & 0.6854 &  678 & 184/234 & (79\%) & 50/234 & (21\%) \\
CT84    &  240 & \texttt{average} & \texttt{chebychev}  & 0.6633 &  239 &  59/82 & (72\%) &  23/82 & (28\%) \\
CM91    &   39 & \texttt{weighted} & \texttt{chebychev} & 0.8129 &   38 &  12/14 & (86\%) &  02/14 & (14\%) \\
ROB97   &  120 & \texttt{average} & \texttt{euclidean}  & 0.7100 &  119 &  39/45 & (87\%) & 06/45 & (13\%) \\
OSTIA1  &  120 & \texttt{average} & \texttt{chebychev}  & 0.7173 &  119 &  36/44 & (82\%) & 08/44 & (18\%) \\
OSTIA2 &  142 & \texttt{average} & \texttt{chebychev}   & 0.6727 &  141 &  30/49 & (61\%) & 19/49 & (39\%) \\
OSTIA3 &  186 & \texttt{average} & \texttt{euclidean}   & 0.6517 &  185 &  31/67 & (46\%) & 36/67 & (54\%) \\
OSTIA4 &   74 & \texttt{average} & \texttt{chebychev}   & 0.7470 &   73 &  14/26 & (53\%) & 12/26 & (46\%) \\
ALL              & 2103 & \texttt{average} & \texttt{cosine}     & 0.6581 & 2102 & 463/742 & (62\%) & 279/742 & (38\%)\\
\midrule
\emph{Overall:} & - & - & - & - & 4194 & 985/1484 & (66\%) & 499/1484 & (34\%)\\
\bottomrule
\end{tabularx}
\end{table}
\vspace{-3em}


\section{Conclusions and Future Works} 
In this paper, we provide evidence that the hierarchical clustering of shape features extracted from the latent subspace of stacked sparse autoencoders (SSAE) is an effective tool for exploring shape similarities for advanced applications in Cultural Heritage science. In particular, we applied this workflow on a newly-introduced database of \emph{ROman COmmonware POTtery} profiles (\texttt{ROCOPOT}), supplying experts with additional comparison tools. 
The most obvious advantage provided by these tools rests in their ability to ease and speed up the processing of a very large number of profiles from different pottery catalogues within a coherent and unified analytical environment. In principle, this could be achieved by pre-sorting the dataset in the way described, presenting the pottery specialist with a selection of most likely matches, thus providing invaluable pointers in the development of a comprehensive typology of Roman commonware. However, this is yet to be achieved, as the complex and partial nature of the profiles makes the clustering problem very significant, especially for big shape variance and the lack of a commonly perceived ground truth. In the near future we envisage two research directions: firstly, we plan to increase the number of profiles available in our database to nearly 5000 objects; secondly, with the help of continuous feedback from specialists, we will focus on meaningful subregions of interest so as to explore if the relevancy of the rim can improve the results of the proposed approach. We expect to not only distinguish between lids and pans, but to unveil more patterns within each single category.

\paragraph{Acknowledgements.} 
\ifdraft
The authors acknowledge the support from the Leverhulme Trust Research Project Grant (RPG-2018-121) ``Unveiling the Invisible - Mathematics for Conservation in Arts and Humanities''. CBS further acknowledges support from the RISE projects CHiPS and NoMADS, the Cantab Capital Institute for the Mathematics of Information and the Alan Turing Institute.
\else
Omitted in the review copy.
\fi

\vspace{-0.75em}
\bibliographystyle{splncs04}
\bibliography{biblio}
\end{document}